\pdfoutput=1
\documentclass{article}

\let\originaladdcontentsline\addcontentsline

\usepackage{iclr2027_conference,times}

\usepackage{amsmath,amsfonts,bm}

\def\eqref#1{equation~\ref{#1}}

\def\1{\bm{1}}

\DeclareMathAlphabet{\mathsfit}{\encodingdefault}{\sfdefault}{m}{sl}
\SetMathAlphabet{\mathsfit}{bold}{\encodingdefault}{\sfdefault}{bx}{n}

\let\addcontentsline\originaladdcontentsline

\usepackage[utf8]{inputenc}
\usepackage[T1]{fontenc}
\usepackage{url}
\usepackage{booktabs}
\usepackage{tabularx}
\usepackage{multirow}
\usepackage{xcolor}
\usepackage{colortbl}
\usepackage{graphicx}
\usepackage{float}
\usepackage{wrapfig}
\usepackage{afterpage}
\usepackage{amsmath}
\usepackage{amsfonts}
\usepackage{nicefrac}
\usepackage{microtype}
\usepackage{xspace}
\usepackage[most]{tcolorbox}%
\usepackage{listings}
\usepackage{wasysym}
\usepackage{amssymb}
\usepackage{titletoc}
\usepackage{enumitem}
\usepackage{hyperref}
\usepackage[capitalise,noabbrev]{cleveref}

\newtcolorbox{promptbox}{%
  breakable,
  enhanced,
  colback=gray!8,
  colframe=gray!40,
  boxrule=0.4pt,
  arc=1pt,
  left=6pt,right=6pt,top=4pt,bottom=4pt,
  before skip=4pt, after skip=4pt,
}

\lstdefinestyle{promptcode}{%
  basicstyle=\ttfamily\scriptsize,
  breaklines=true,
  breakatwhitespace=false,
  columns=fullflexible,
  keepspaces=true,
  showstringspaces=false,
  aboveskip=0pt,
  belowskip=0pt,
}
\lstnewenvironment{promptlisting}{\lstset{style=promptcode}}{}

\definecolor{rolePivot}{HTML}{B22222}
\definecolor{traceGray}{HTML}{6A6A6A}
\lstdefinestyle{trajcode}{%
  style=promptcode,
  basicstyle=\ttfamily\scriptsize\color{traceGray},
  moredelim=[is][\bfseries\color{black}]{|B|}{|B|},
  moredelim=[is][\itshape\color{black}]{|EL|}{|EL|},
  moredelim=[is][\color{rolePivot}\bfseries]{|HL|}{|HL|},
  moredelim=[is][\color{black}]{|N|}{|N|},
}
\lstnewenvironment{trajlisting}{\lstset{style=trajcode}}{}
\newtcolorbox{trajbox}[1]{%
  breakable, enhanced, colback=gray!8, colframe=gray!45, boxrule=0.4pt, arc=1pt,
  left=6pt,right=6pt,top=3pt,bottom=3pt, before skip=6pt, after skip=6pt,
  fonttitle=\footnotesize\bfseries, coltitle=black, colbacktitle=gray!20, colupper=traceGray, title={#1},
}

\newcolumntype{L}[1]{>{\raggedright\arraybackslash}p{#1}}
\newcolumntype{C}[1]{>{\centering\arraybackslash}p{#1}}
\newcolumntype{Y}{>{\raggedright\arraybackslash}X}
\newcolumntype{Z}{>{\centering\arraybackslash}X}

\definecolor{ars}{HTML}{AA66DD}

\newcommand{\benchname}{\textsc{SciLaws-Bench}\xspace}
\newcommand{\realname}{\textsc{SciLaws-Real}\xspace}
\newcommand{\parallelname}{\textsc{SciLaws-Parallel}\xspace}

\newcommand{\answerTODO}[1][]{\textcolor{red}{\bf [TODO]}}
\newcommand{\justificationTODO}[1][]{\textcolor{red}{\bf [TODO]}}

\title{Can LLMs Discover Scientific Laws in Real and Parallel Worlds?}

\author{\textbf{Yiming Huang}\textsuperscript{1,}\thanks{Equal contribution.} \quad
\textbf{Ziche Liu}\textsuperscript{1,}\footnotemark[1] \quad
\textbf{Junxia Cui}\textsuperscript{1} \quad
\textbf{Zhuohang Wu}\textsuperscript{2} \quad
\textbf{Yiqian Wang}\textsuperscript{3}\\%
\textbf{Xinkai Zou}\textsuperscript{1} \quad
\textbf{Lingjun Mao}\textsuperscript{1} \quad
\textbf{Nan Huang}\textsuperscript{1} \quad
\textbf{Naicheng Yu}\textsuperscript{1}\\%
\textbf{Kaijie Zhu}\textsuperscript{4} \quad
\textbf{Yue Ma}\textsuperscript{1} \quad
\textbf{Kun Zhou}\textsuperscript{1} \quad
\textbf{Letian Peng}\textsuperscript{1,}\thanks{Corresponding authors.} \quad
\textbf{Jingbo Shang}\textsuperscript{1,}\footnotemark[2]\\%
\textsuperscript{1}University of California San Diego \quad
\textsuperscript{2}University of California, Irvine\\%
\textsuperscript{3}Cushing Academy \quad
\textsuperscript{4}University of California, Santa Barbara\\%
\texttt{\{yih112,zil199,jshang,lepeng\}@ucsd.edu}
}

\iclrfinalcopy

\begin{document}

\maketitle
\lhead{}
\renewcommand{\headrulewidth}{0pt}

\begin{abstract}

Scientific law discovery has long been central to scientific progress, proceeding through iterative cycles of generating hypotheses, testing them against empirical evidence, and refining them under scientific constraints.
As large language models (LLMs) become increasingly involved in scientific research, whether they can discover scientific laws and how to evaluate this ability remain open questions.
A central evaluation challenge is to move beyond familiar published equations while keeping discovery tasks grounded in scientific data and constraints.
We introduce \textbf{\benchname{}}, a curated collection of scientific task packages grounded in the source literature, each linking a scientific problem, supporting data, published reference equations, and scientific-validity rubrics.
Through agent-assisted curation and human verification, we assemble 118 problems spanning six disciplines, drawing on 381 papers, 291 candidate laws, and roughly 8M data points.
Each problem supports two complementary evaluation settings.
\textbf{\realname{} }uses fixed scientific data to evaluate proposed laws for held-out predictive fit and scientific validity.
\textbf{\parallelname{}} evaluates recovery of a newly synthesized structural variant of a published equation through active queries to a simulator calibrated to the source data.
Our evaluation reveals three limitations: good predictive fit need not imply scientific validity, recovering a published formula does not establish recovery of its new structural terms, and candidate selection remains a bottleneck in scientific law discovery. \textbf{Project page:} \url{https://yiyihum.github.io/SciLaws-Bench}.
\end{abstract}

\afterpage{%

\begin{figure}[t!]
\centering
\includegraphics[width=\linewidth]{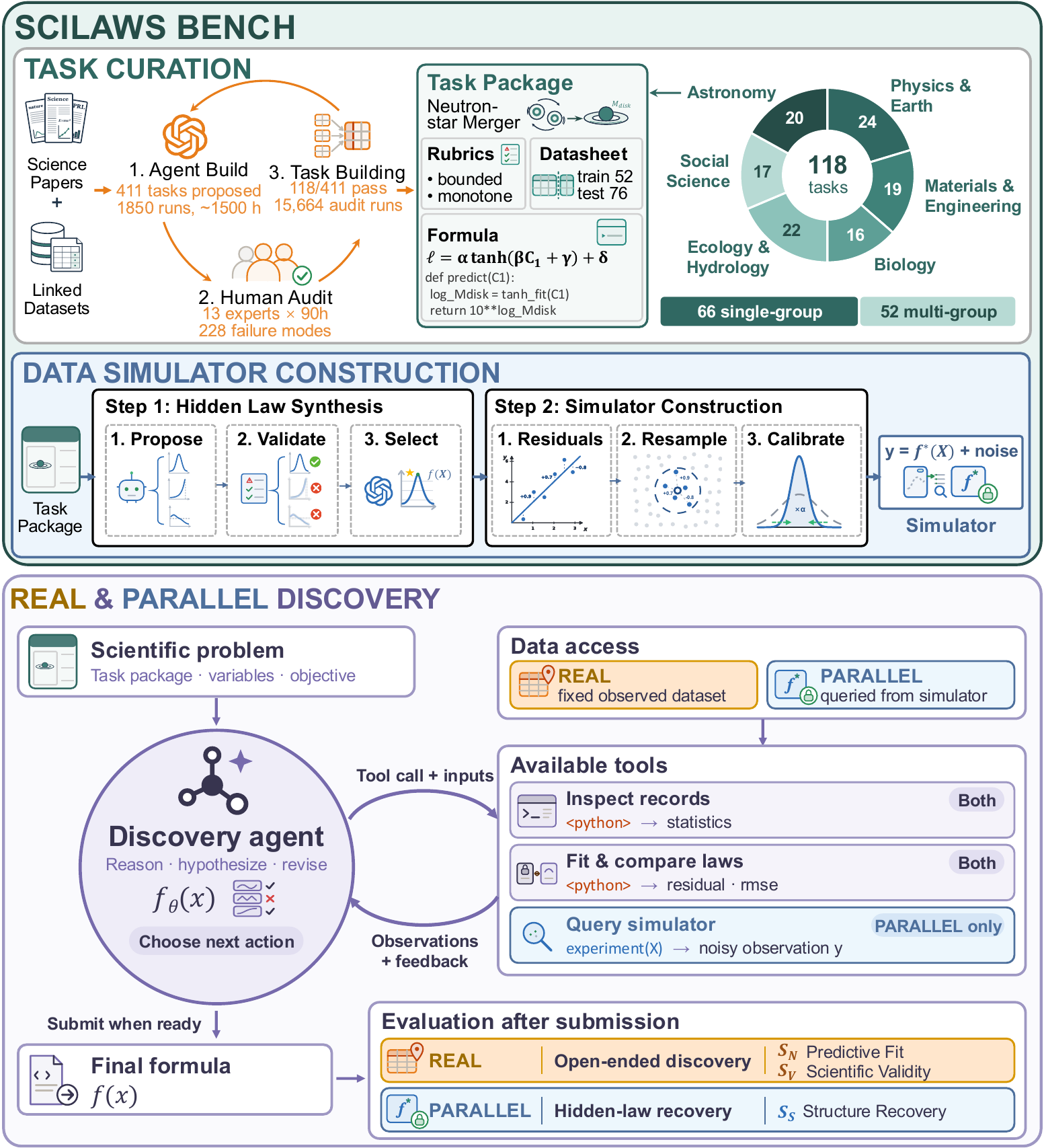}
\caption{Overview of \benchname{}.}
\label{fig:main}
\end{figure}
}

\section{Introduction}
\label{sec:introduction}
From Kepler's planetary laws to Hubble's law, scientific laws have provided compact mathematical descriptions of imperfect measurements. Equation discovery, often formalized as symbolic regression, seeks interpretable mathematical expressions that explain the observed data \citep{langley1987scientificdiscovery, schmidt2009eureqa}. Scientific law discovery requires more than goodness of fit or syntactic simplicity. Proposed laws must also respect physical constraints, remain well behaved in the relevant regime, and retain scientific meaning in context.
As LLMs take on a larger role in scientific discovery~\citep{shojaee2025llmsr,lu2024aiscientist}, evaluating whether they can meet these requirements becomes essential.

Existing evaluations use several approaches to study equation recovery and scientific reasoning.
Benchmarks like AI Feynman \citep{udrescu2020aifeynman} evaluate the recovery of canonical textbook equations (e.g., $E=\hbar\omega$ and $\theta_1=\arcsin(n\sin\theta_2)$), for which prior familiarity can contribute to recovery.
Counterfactual benchmarks such as NewtonBench \citep{zheng2026newtonbench} modify familiar equations to create controlled targets (e.g., replacing \(F=Gm_1m_2/r^2\) with \(F=G'm_1m_2/r^{1.5}\)), while DiscoveryWorld \citep{jansen2024discoveryworld} uses simulated environments to study scientific reasoning and experimentation.
The challenge is to evaluate discovery beyond familiar equations while keeping tasks grounded in scientific data and domain-specific constraints.

To address this challenge, we introduce\textbf{ \benchname{} }(\cref{fig:main}), a curated collection of scientific task packages grounded in the source literature, each including a scientific problem, supporting data, published reference equations, and scientific-validity rubrics.
It comprises 118 problems from ongoing scientific research rather than standard textbook examples, drawing on 381 papers, 291 candidate laws, and roughly 8M data points across six disciplines.
Constructing these task packages requires linking evidence across papers and datasets, standardizing heterogeneous data, implementing published formulas, and translating scientific constraints into explicit validation criteria.
We combine agent-assisted curation with domain-expert guidance and human verification of critical variable mappings, formula implementations, data splits, and rubric items.
Together, this curation and verification turn scattered scientific evidence into task packages that support both law discovery and evaluation against explicit scientific criteria.
Building on this foundation, we instantiate two complementary evaluation settings: \textbf{\realname{}} fixes the existing data and leaves the candidate law open, whereas\textbf{ \parallelname{} }fixes a synthesized generating equation and lets models query new data from its simulator.

\textbf{\realname{}} asks models to discover closed-form laws from existing scientific data (Section~\ref{sec:real_setting}).
The goal is to find laws that better explain held-out observations than published formulas while respecting the physical and other scientific constraints of each problem.
These data include experimental measurements, observations, and outputs of scientific simulations.
Published formulas serve as reference baselines rather than recovery targets, and proposals are evaluated separately for held-out predictive fit and source-grounded scientific validity.

\textbf{\parallelname{}} introduces controlled conditions for the same problem by constructing a queryable simulator with a fixed hidden generating equation, while preserving the problem's scientific context (Section~\ref{sec:parallel_setting}).
Its generating equation is a newly synthesized structural variant of a published formula, subject to the problem's scientific constraints (e.g., adding the distance-dependent anelastic attenuation term to a published ground-motion law).
The coefficients and residual noise are calibrated to the corresponding source data.
Starting without observations, models choose simulator queries and infer the fixed hidden equation from the returned responses, enabling evaluation of structural recovery against a target known by construction.

On \benchname{}, we evaluate fourteen LLMs spanning leading proprietary and open-weight families, including GPT-6, Claude Opus 5.5, Gemini 3.5 Flash, and Kimi K3. All models are evaluated under the same ReAct-style agent framework, with access to a Python sandbox and, in \parallelname{}, a fixed-budget experimentation interface, for up to 30 interaction turns per trial.
A closer analysis of their trajectories further reveals distinct discovery styles: some models emphasize breadth, screening many candidate laws in a few turns, while others emphasize depth, refining a few candidates over many turns.
Our analysis reveals three central findings.

\begin{enumerate}[leftmargin=*, nosep]
\item Models struggle to balance predictive fit and scientific validity. For example, a model's best-fitting formula on a nuclear-physics task introduces a nonexistent resonance peak. It therefore fails our validity checks.

\item Models use memory to reproduce known laws, but struggle to discover novel structure. Reference-beating is more common on tasks with low recall, while full-law recovery varies substantially across models.

\item Models generate better laws than they can reliably select. Scoring the intermediate candidate laws within trajectories shows that weaker models propose laws as good as the leading models' but submit worse ones. Best-of-\(N\) search exposes the same gap across trajectories, with self-selection capturing only a small fraction of the available gains.
\end{enumerate}

Overall, current LLMs show some ability to discover scientific laws, but do not yet do so reliably.
Self-evaluation and candidate selection remain a bottleneck in scientific law discovery, as agents do not consistently select the best laws available in their candidate pools.

\section{\benchname}

\subsection{Benchmark Overview}
\label{sec:benchmark_overview}

As illustrated in Figure~\ref{fig:main}, we curate linked papers and existing scientific data to construct fixed-data \realname{} tasks, then build corresponding \parallelname{} calibrated simulator  for evaluation using synthesized hidden equations derived from published forms. \benchname{} comprises 118 problems across six disciplines, 291 candidate laws, 381 scientific papers, and roughly 8M data points.
To our knowledge, \benchname{} is the first scientific law discovery benchmark to include both single-group and multi-group problems.
Single-group problems use one global law and parameter set per task, whereas multi-group problems share the same functional form across groups while allowing some parameter values to vary.
The benchmark contains 66 single-group and 52 multi-group problems.
Together, these design choices enable a comprehensive evaluation of scientific law discovery, from fitting a single system to identifying laws shared across groups, using either existing data or data collected through active experimentation.

\subsection{Scientific Task Curation}
\label{sec:problem_selection}

\paragraph{Problem selection.}
We collect linked paper–dataset pairs across six scientific domains with LLM-assisted search and initial screening, followed by human verification of critical eligibility decisions. We focus on scientifically active, data-driven problems in collaboration with domain experts, prioritizing topics where published laws leave room for improvement. Each retained problem must satisfy three criteria: (1) the paper presents an established closed-form law with enough validity constraints; (2) the dataset is publicly available and contains sufficient observations for reliable evaluation; and (3) jointly, we exclude target leakage and require at least one published formula to outperform a naive baseline.

\begin{wrapfigure}{r}{0.42\linewidth}
\centering
\includegraphics[width=\linewidth]{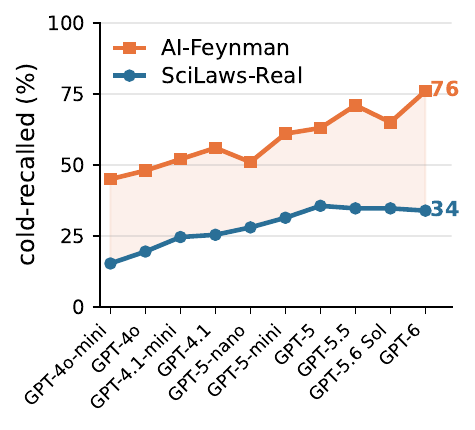}
\caption{Memorization comparison.}
\label{fig:memorization_feynman_vs_realsr}
\end{wrapfigure}

\paragraph{Task package construction.}
After filtering, papers are converted to Markdown and relevant task fields are extracted, while associated data tables are merged and standardized for task construction. From curated papers and datasets, we build task metadata, executable reference formulas in Python, cleaned data splits, and a frozen source-grounded validity rubric. Human reviewers verify critical variable mappings, formula implementations, split rationales, and rubric items against the cited sources. These packages provide the common foundation for both \realname{} and \parallelname{} (worked example in Figure~\ref{fig:pipeline_detailed}).

\paragraph{Memorization audit.}
To assess familiarity with the selected reference equations, we conduct a matched ten-model cold-recall audit (Section~\ref{sec:memorization} ). Recall is substantially lower for \realname{} reference forms than for AI-Feynman formulas~\citep{udrescu2020aifeynman} (\cref{fig:memorization_feynman_vs_realsr}), indicating that the selected tasks are less readily addressed through direct formula recall.

\subsection{\realname{}}
\label{sec:real_setting}

Scientific law discovery often proceeds from existing scientific data, including experimental measurements, observations, and outputs of scientific simulations. \realname{} evaluates whether an LLM-guided system can use these fixed data and their scientific context to propose closed-form laws that surpass published references. The reference formulas provide comparison baselines without specifying a unique target equation.

\paragraph{Held-out splits.}
Grounded in the source literature, single-group tasks use task-specific splits for interpolation, input-range extrapolation, temporal transfer, or cross-condition generalization.
Multi-group tasks hold out entire groups and divide each test group into a calibration subset to fit per-group parameters and a disjoint evaluation subset to evaluate the shared law.
For example, the multi-group optical-dispersion task trains on 11 crystal materials and holds out three unseen materials: CdSe, TiO\(_2\), and ZnWO\(_4\).
For each held-out material, short-wavelength measurements calibrate the per-group parameters and long-wavelength measurements evaluate the shared law.
Task scores are averaged equally across the discriminating test groups (Appendix~\ref{app:eval_numeric}).

\paragraph{Evaluation.}
\label{sec:evaluation}
For a task \(t\), \realname{} reports numeric fit \(S_N(t)\) and scientific validity \(S_V(t)\), whose task-set averages are \(S_N\) and \(S_V\).
We report them separately because a law can fit held-out records while violating paper-grounded scientific requirements.

Each task uses one primary metric from \{\texttt{rmse}, \texttt{smape}, \texttt{log\_mae}\} (lower is better) or \texttt{r2} (higher is better).
The raw metric value of the best-performing formula defines \(m_{\mathrm{ref}}\).
A submission with metric value \(m_{\mathrm{sub}}\) is normalized as
\begin{equation}
S_N(t) =
\begin{cases}
\operatorname{clip}\!\left(1-\frac{1}{2}\frac{m_{\mathrm{sub}}}
{m_{\mathrm{ref}}},\,0,\,1\right),
& \texttt{rmse},\ \texttt{smape},\ \texttt{log\_mae}\;,\\
\operatorname{clip}\!\left(0.5+\frac{1}{2}
\frac{m_{\mathrm{sub}}-m_{\mathrm{ref}}}
{1-m_{\mathrm{ref}}},\,0,\,1\right),
& \texttt{r2}\;.
\end{cases}
\end{equation}
The clipping operator bounds \(S_N(t)\) to \([0,1]\).
The strongest published formula scores \(0.5\), and a perfect predictor scores \(1.0\).

To compute \(S_V(t)\), a code-enabled judge evaluates each item in the frozen task rubric through deterministic probes and source inspection.
A global anti-hacking item rejects submissions that move degrees of freedom outside the declared law, such as lookup tables or excessive fitted constants.
Let \(a_t=1\) if the submission passes the anti-hacking check and \(a_t=0\) otherwise, and let \(M_t\) denote the number of satisfied items among all \(N_t\) rubric items.
\begin{equation}
S_V(t)=a_t\,\frac{M_t}{N_t}.
\end{equation}

\subsection{\parallelname{}}
\label{sec:parallel_setting}

Scientific discovery often involves deciding what to measure as well as interpreting existing scientific data.
\parallelname{} evaluates active structural recovery through simulated experiments in a corresponding generating environment constructed for each curated problem. It preserves from \realname{} the scientific context, admissible input ranges, validity constraints, and group structure, while providing a residual-calibrated simulator with a fixed hidden generating equation. The hidden target is a newly synthesized structural variant of a published formula.
Starting without observations, models select inputs (and groups for multi-group tasks) within a task-specific query budget and infer the hidden equation from simulator responses.

\paragraph{Hidden law synthesis.} 
Starting from the selected published baseline, multiple agents independently propose structural variants. Candidates go through an initial execution and fit check, followed by scientific-validity and integrity screening. Survivors will be evaluated by a final judge based on simplicity, physical plausibility and source-consistency, yielding one winner as the hidden law \(F_t\).

\begin{figure}[h]
\centering
\includegraphics[width=\linewidth]{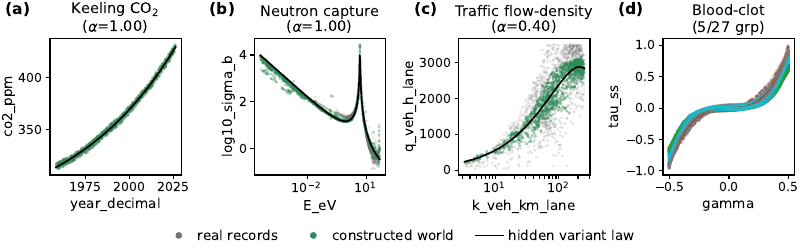}
\vspace{-10pt}
\caption{\parallelname{} worlds and the corresponding source data.
\textbf{(a)--(c)} show single-group tasks, and \textbf{(d)} shows a multi-group task with one shared functional form and group-specific parameter values.
\(\alpha\) denotes the residual-noise calibration factor, where \(1.00\) retains the debiased residual scale.}
\label{fig:sim_worlds}
\end{figure}

\paragraph{Simulator construction.} 
For a hidden law, parameters are fitted to the source data globally for single-group tasks and separately within each group for multi-group tasks.
Residuals are then calculated in a linear or logarithmic residual space.
We remove estimated local trends from the residuals and resample them using a k-NN bootstrap.
We calibrate the residual scale relative to the hidden-law signal to limit noise without imposing a common noise distribution across tasks.
Finally, the resulting simulator outputs \(y = F_t(X) + \alpha\ \cdot\ \mathrm{resample}(X)\) in the task's residual space, with a valid input range and query budget (representative worlds in Figure~\ref{fig:sim_worlds}).

\paragraph{Evaluation.}
\parallelname{} measures recovery of hidden structure rather than predictive fit.
A code-enabled judge compares each submission with the hidden law (protocol in Appendix~\ref{sec:eval_appendix}).
The five \(S_S\) levels are \(0\) (invalid or unrelated), \(0.25\) (relevant variables or trends), \(0.5\) (published base form), \(0.75\) (base form with most added terms), and \(1\) (complete hidden structure).

\section{Experiments and Analysis}
\label{sec:analysis}

\subsection{Setup}
\label{sec:experiments_setup}

\paragraph{Protocol and Models.}
All base models use a shared ReAct-style agent framework with up to 30 interaction turns~\citep{yao2023react}.
The agent uses Python to analyze observations, fit parameters, and compare candidate formulas.
Active experimentation is available in \parallelname{}, where the agent selects inputs and, for multi-group tasks, groups to query within a task-specific budget.
Each run ends with a formula submitted as Python code.
We evaluate models from the GPT, Claude, Gemini, DeepSeek, Qwen, Kimi, and GLM families (Table~\ref{tab:main_results}).
Models with configurable reasoning effort use the medium setting.

\paragraph{Judges and Human Validation.}
GPT-5.4-mini with xhigh reasoning effort evaluates scientific validity and structural recovery, while GPT-4.1 judges memorization.
On a stratified sample of 135 items per judge, against majority labels from five domain experts, Cohen's \(\kappa\) is \(0.82\) for validity, \(0.84\) for structure (quadratically weighted), and \(0.77\) for memorization.
Each falls within the corresponding pairwise human--human range (Appendix~\ref{app:human_alignment}).

\begin{wrapfigure}{r}{0.45\linewidth}
\vspace{-15pt}
\centering
\includegraphics[width=\linewidth]{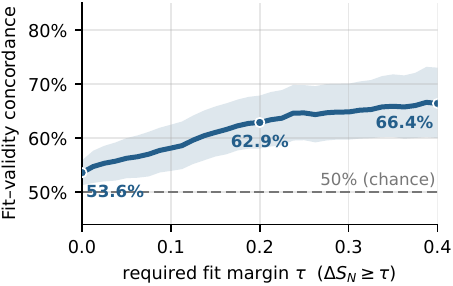}
\caption{\textbf{Fit--validity concordance.} Shading marks 95\% task-bootstrap confidence intervals.}
\label{fig:coupling_domains}
\vspace{-25pt}
\end{wrapfigure}

\subsection{Model Performance}
\label{sec:main_results}

Table~\ref{tab:main_results} summarizes model performance on predictive fit, scientific validity, and hidden-law recovery.
Compared with GPT-5.5 and Claude Opus 4.8, GPT-6 and Claude Opus 5.5 improve \(S_S\) by 14.2 and 8.7 points, respectively, while their \(S_V\) scores increase by 1.1 and 0.5 points.
Paired task-bootstrap comparisons support GPT-6's improvement over GPT-5.5 in structural recovery, but do not establish corresponding improvements in predictive fit or scientific validity (Appendix~\ref{app:paired_model_comparison}).

\begin{table}[!t]
\centering
\footnotesize
\setlength{\tabcolsep}{1.8pt}
\definecolor{rankc}{HTML}{4A7EBB}
\renewcommand{\arraystretch}{1.12}
\caption{Results on \benchname{} (\%). All: per-task mean over all tasks; bold/underline: best/second-best value; darker shading: better rank. Submit: share of the 236 runs (both settings) returning a runnable formula; Hack: share of \realname{} tasks failing the anti-hacking check.}
\label{tab:main_results}
\begin{tabularx}{\linewidth}{@{}L{0.2\linewidth} *{9}{Z} C{0.066\linewidth} C{0.064\linewidth}@{}}
\toprule
& \multicolumn{3}{c}{\realname{} \(S_N\)} & \multicolumn{3}{c}{\realname{} \(S_V\)} & \multicolumn{3}{c}{\parallelname{} \(S_S\)} & \shortstack{Submit\\$\uparrow$} & \shortstack{Hack\\$\downarrow$} \\
\cmidrule(lr){2-4}\cmidrule(lr){5-7}\cmidrule(lr){8-10}
Base Model & Single & Multi & All & Single & Multi & All & Single & Multi & All & (\%) & (\%) \\
\midrule
GPT-6 & \cellcolor{rankc!40}\textbf{52.67} & \cellcolor{rankc!40}\textbf{49.62} & \cellcolor{rankc!40}\textbf{51.33} & \cellcolor{rankc!40}\textbf{88.21} & \cellcolor{rankc!37}\underline{76.19} & \cellcolor{rankc!40}\textbf{82.91} & \cellcolor{rankc!40}\textbf{70.83} & \cellcolor{rankc!40}\textbf{74.52} & \cellcolor{rankc!40}\textbf{72.46} & \cellcolor{rankc!34}98.3 & \cellcolor{rankc!37}\underline{0.8} \\
GPT-5.5 & \cellcolor{rankc!37}\underline{52.44} & \cellcolor{rankc!34}47.15 & \cellcolor{rankc!37}\underline{50.11} & \cellcolor{rankc!31}87.79 & \cellcolor{rankc!31}74.29 & \cellcolor{rankc!37}\underline{81.84} & \cellcolor{rankc!37}\underline{60.23} & \cellcolor{rankc!34}55.77 & \cellcolor{rankc!34}58.26 & \cellcolor{rankc!37}\underline{98.7} & \cellcolor{rankc!28}1.7 \\
Claude Opus 5.5 & \cellcolor{rankc!31}49.21 & \cellcolor{rankc!37}\underline{48.45} & \cellcolor{rankc!34}48.87 & \cellcolor{rankc!37}\underline{88.04} & \cellcolor{rankc!22}71.55 & \cellcolor{rankc!28}80.77 & \cellcolor{rankc!37}\underline{60.23} & \cellcolor{rankc!37}\underline{56.73} & \cellcolor{rankc!37}\underline{58.69} & \cellcolor{rankc!22}97.5 & \cellcolor{rankc!22}2.5 \\
Kimi K3 & \cellcolor{rankc!28}48.11 & \cellcolor{rankc!31}46.40 & \cellcolor{rankc!31}47.36 & \cellcolor{rankc!28}85.81 & \cellcolor{rankc!15}69.69 & \cellcolor{rankc!22}78.71 & \cellcolor{rankc!12}48.86 & \cellcolor{rankc!22}46.15 & \cellcolor{rankc!12}47.67 & \cellcolor{rankc!34}98.3 & \cellcolor{rankc!3}6.8 \\
Gemini 3.5 Flash & \cellcolor{rankc!25}46.67 & \cellcolor{rankc!25}45.34 & \cellcolor{rankc!28}46.08 & \cellcolor{rankc!22}84.69 & \cellcolor{rankc!3}61.86 & \cellcolor{rankc!15}74.63 & \cellcolor{rankc!25}54.17 & \cellcolor{rankc!15}45.67 & \cellcolor{rankc!28}50.42 & \cellcolor{rankc!15}96.2 & 9.3 \\
GLM-5.2 & \cellcolor{rankc!34}49.29 & \cellcolor{rankc!9}41.46 & \cellcolor{rankc!25}45.84 & \cellcolor{rankc!15}81.75 & \cellcolor{rankc!9}64.58 & \cellcolor{rankc!12}74.19 & \cellcolor{rankc!31}56.44 & \cellcolor{rankc!25}46.63 & \cellcolor{rankc!31}52.12 & \cellcolor{rankc!9}95.3 & \cellcolor{rankc!6}5.1 \\
Claude Opus 4.8 & \cellcolor{rankc!22}45.71 & \cellcolor{rankc!18}43.96 & \cellcolor{rankc!22}44.94 & \cellcolor{rankc!34}88.03 & \cellcolor{rankc!18}70.35 & \cellcolor{rankc!25}80.24 & \cellcolor{rankc!22}53.03 & \cellcolor{rankc!22}46.15 & \cellcolor{rankc!22}50.00 & \cellcolor{rankc!18}96.6 & \cellcolor{rankc!37}\underline{0.8} \\
DeepSeek-V4 Pro & \cellcolor{rankc!12}42.21 & \cellcolor{rankc!28}46.18 & \cellcolor{rankc!18}43.96 & \cellcolor{rankc!18}83.75 & \cellcolor{rankc!40}\textbf{78.19} & \cellcolor{rankc!34}81.30 & \cellcolor{rankc!18}51.14 & \cellcolor{rankc!31}48.08 & \cellcolor{rankc!18}49.79 & \cellcolor{rankc!40}\textbf{99.2} & \cellcolor{rankc!37}\underline{0.8} \\
Qwen3.7-Max & \cellcolor{rankc!18}43.39 & \cellcolor{rankc!22}44.33 & \cellcolor{rankc!15}43.80 & \cellcolor{rankc!3}73.39 & \cellcolor{rankc!25}71.72 & \cellcolor{rankc!6}72.66 & \cellcolor{rankc!28}54.55 & \cellcolor{rankc!12}45.19 & \cellcolor{rankc!28}50.42 & \cellcolor{rankc!6}93.2 & \cellcolor{rankc!18}3.4 \\
GPT-5-mini & \cellcolor{rankc!9}39.79 & \cellcolor{rankc!12}41.82 & \cellcolor{rankc!12}40.69 & \cellcolor{rankc!9}79.39 & \cellcolor{rankc!28}74.03 & \cellcolor{rankc!18}77.03 & \cellcolor{rankc!6}44.70 & \cellcolor{rankc!9}44.71 & \cellcolor{rankc!6}44.70 & \cellcolor{rankc!25}97.9 & \cellcolor{rankc!18}3.4 \\
DeepSeek-V4.1 Flash & \cellcolor{rankc!15}42.83 & \cellcolor{rankc!3}37.76 & \cellcolor{rankc!9}40.59 & \cellcolor{rankc!12}81.69 & \cellcolor{rankc!6}64.41 & \cellcolor{rankc!9}74.08 & \cellcolor{rankc!15}50.00 & \cellcolor{rankc!6}42.31 & \cellcolor{rankc!9}46.61 & 90.3 & \cellcolor{rankc!40}\textbf{0.0} \\
GPT-5.6 Sol & \cellcolor{rankc!6}37.24 & \cellcolor{rankc!15}43.75 & \cellcolor{rankc!6}40.11 & \cellcolor{rankc!25}84.90 & \cellcolor{rankc!34}75.90 & \cellcolor{rankc!31}80.94 & \cellcolor{rankc!9}48.11 & \cellcolor{rankc!31}48.08 & \cellcolor{rankc!15}48.09 & \cellcolor{rankc!34}98.3 & \cellcolor{rankc!12}4.2 \\
GPT-5.4-mini & \cellcolor{rankc!3}36.62 & \cellcolor{rankc!6}37.86 & \cellcolor{rankc!3}37.17 & \cellcolor{rankc!6}76.70 & \cellcolor{rankc!12}67.06 & \cellcolor{rankc!3}72.45 & \cellcolor{rankc!3}43.18 & \cellcolor{rankc!3}41.83 & \cellcolor{rankc!3}42.58 & \cellcolor{rankc!15}96.2 & \cellcolor{rankc!12}4.2 \\
GPT-4o-mini & 27.27 & 12.12 & 20.59 & 68.17 & 40.60 & 56.02 & 34.47 & 32.21 & 33.47 & \cellcolor{rankc!3}91.1 & \cellcolor{rankc!28}1.7 \\

\bottomrule
\end{tabularx}
\end{table}

\paragraph{Better fit does not reliably imply higher validity or structural recovery.}
\label{sec:numeric_vs_validity}
Within the same task, fit and validity agree in only \(53.6\%\) of \(9{,}279\) unequal-fit model pairs, counting validity ties as half-agreement.
Concordance rises to \(66.4\%\) at a fit margin of \(\Delta S_N\geq0.4\) (Figure~\ref{fig:coupling_domains}; Appendix~\ref{app:fit_validity_concordance}).
Across the two settings, mean \(S_N\) and \(S_S\) correlate across models (\(r=0.83\)), but their task-level correlation within GPT-6 is only \(r=0.06\), using the main table's strict scoring convention.
On the Keeling task, GPT-5.5 exceeds the published reference in \realname{} but recovers only the trend in \parallelname{}.
On Sellmeier, it falls below the reference in \realname{} yet fully recovers the hidden law in \parallelname{} (Appendices~\ref{app:case_keeling} and~\ref{app:case_sellmeier}).
Aggregate improvements therefore do not make predictive fit a reliable proxy for scientific validity or structural recovery on individual tasks.

\subsection{How Memorization Shapes Scientific Discovery}
\label{sec:memorization}

Memory is a central challenge in evaluating scientific agents, since a formula reproduced from memory may be indistinguishable from one inferred from task evidence \citep{xu2024benchmarkdatacontamination,li2023taskcontamination}.
We first ask how readily models can reproduce a published formula from a restricted description of the scientific problem, before seeing any task data.
We use a task-level memorization audit to examine how recall relates to surpassing published baselines and recovering structure absent from the published formula.

\paragraph{Memorization audit.}
The audit tests whether a model can reproduce the best published reference formula from the input and output variable specification alone.
Following extraction-based measurement of memorization~\citep{carlini2021extracting,hayes2025probabilistic}, we count a formula as cold-recalled by a model when at least three of five generated Python implementations structurally match the reference, as assessed by a calibrated LLM judge.
This partitions the 118 tasks into Canon (11), recalled by all fourteen models, Moat (46), recalled by none, and Mid (61), recalled by some but not all models.
Universal recall is limited to \(9.3\%\) of tasks, whereas \(39.0\%\) form a discovery moat under this audit.
Even Claude Opus 5.5, which has the highest observed recall rate, reproduces only \(44.1\%\) of the reference formulas.
Appendix~\ref{app:memorization} provides the judging protocol and per-model results.

\begin{figure}[!b]
\centering
\includegraphics[width=\linewidth]{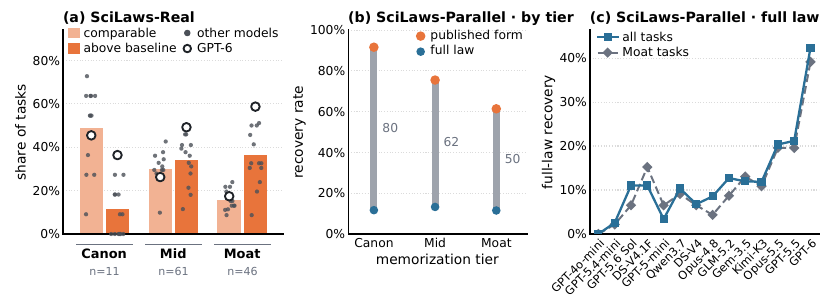}
\caption{\textbf{(a)} Tasks above or near the best baseline by recall tier; bars are model means. \textbf{(b)} Published-form vs.\ full-law recovery by tier. \textbf{(c)} Full-law recovery per model, all tasks vs.\ Moat.}
\label{fig:mem_real_parallel}
\end{figure}

\paragraph{On \realname{}, exceeding the reference is more common on Moat tasks.}
Figure~\ref{fig:mem_real_parallel}(a) distinguishes solutions that exceed the reference (\(S_N>0.52\)) from those that perform comparably (\(|S_N-0.5|\leq0.02\)).
Canon tasks more often yield reference-level fits, whereas Moat tasks more often yield improvements over the published baseline.
Averaged across models, the above-reference share rises from \(12\%\) on Canon to \(36\%\) on Moat, while the reference-level share falls from \(49\%\) to \(16\%\).
The same contrast holds for GPT-6, whose above-reference share increases from \(36\%\) to \(59\%\) and reference-level share decreases from \(45\%\) to \(17\%\).
For example, GPT-5.5 recalls the Somigliana formula on the Canon gravity task and ties the reference, but fits a new form on the Moat chiller-efficiency task that exceeds it (\(S_N=0.81\)).
Exceeding a published baseline is consistent with useful data-driven fitting, but does not by itself establish recovery of the underlying scientific mechanism.

\paragraph{On \parallelname{}, recall is more closely associated with published-form recovery than full-law recovery.}
We distinguish recovery of the published form (\(S_S\geq0.5\)) from recovery of the full synthesized law (\(S_S=1\)).
Across cold-recall tiers, published-form recovery falls from \(92\%\) on Canon to \(61\%\) on Moat, whereas full-law recovery remains within \(12\%\)--\(13\%\) across tiers (Figure~\ref{fig:mem_real_parallel}(b)).
Full-law recovery instead varies substantially across models, from \(0\%\) for GPT-4o-mini to \(42\%\) for GPT-6, which also reaches \(39\%\) on Moat tasks (Figure~\ref{fig:mem_real_parallel}(c)).
The Beer--Lambert task illustrates this distinction: all fourteen models recover the published form, but none recovers the full hidden law.
Reproducing a familiar formula therefore does not establish recovery of its synthesized structural variant.

\begin{figure}[!h]
\centering
\includegraphics[width=0.85\linewidth]{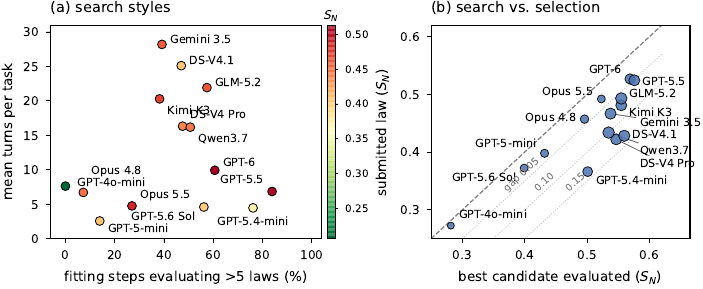}
\caption{\textbf{(a)} Search breadth and interaction length, coloured by numeric performance. \textbf{(b)} Best-candidate versus final-submission performance, with marker size indicating candidate count.}
\label{fig:discovery_process}
\end{figure}

\begin{figure}[!h]
\centering
\includegraphics[width=0.94\linewidth]{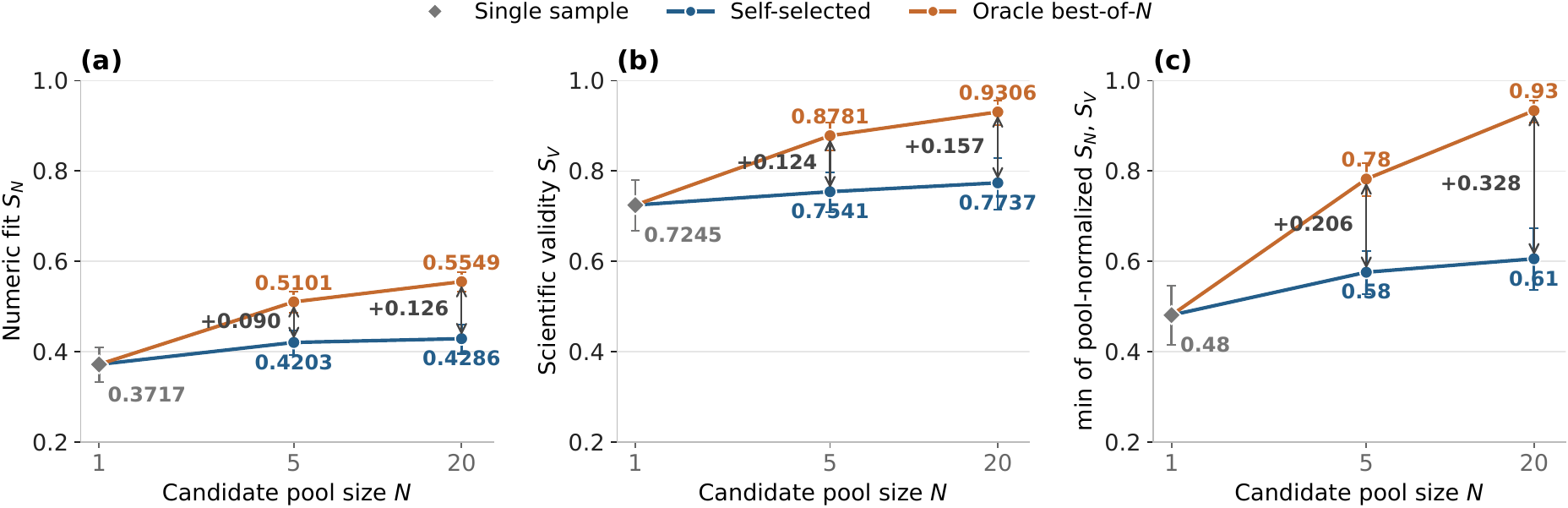}
\caption{Self-selected vs.\ oracle best-of-\(N\) for \textbf{(a)} \(S_N\), \textbf{(b)} \(S_V\), and \textbf{(c)} joint balance (the worse of the two pool-normalized scores). Error bars are 95\% task-bootstrap confidence intervals.}
\label{fig:repeated_sampling}
\end{figure}

\subsection{How Agents Discover: Search and Selection}
\label{sec:repeated_sampling}

Finding a promising law and selecting it as the final answer are distinct challenges.
We study both: within single-group \realname{} trajectories, by scoring every executable intermediate candidate on held-out data against the final submission (Appendix~\ref{app:replay_protocol}), and across independent trajectories, with a best-of-\(N\) experiment that also considers scientific validity.

\paragraph{Models adopt distinct search styles.}
Figure~\ref{fig:discovery_process}(a) characterizes search along two dimensions: interaction length and the fraction of fitting steps that evaluate more than five candidates.
GPT-5.5 and GPT-6 frequently compare multiple candidates within a fitting step, whereas Gemini~3.5~Flash uses more interaction turns.
Search styles also change within model families: Claude Opus~5.5 compares more than five candidates more frequently than Opus~4.8, while using fewer turns.

\paragraph{Within a trajectory, models generate better-fitting laws than they select.}
Figure~\ref{fig:discovery_process}(b) compares the best candidate in each trajectory, including the final submission, with the submitted law.
Several models have comparable best-candidate scores but differ substantially in their final performance.
The mean gap is 0.14 in \(S_N\) for GPT-5.4-mini, compared with 0.03--0.04 for GPT-6 and Claude Opus~5.5.
This gap suggests that finding a well-fitting candidate does not ensure that it becomes the final answer.
The best-candidate score is an offline diagnostic based on held-out data, not a selection signal available to the agent.
Because this comparison measures numeric fit alone, it does not establish whether a higher-scoring candidate also satisfies the task's scientific constraints.

\paragraph{Best-of-\texorpdfstring{\(N\)}{N} search misses gains on both metrics.}
We next study GPT-5.4-mini on all 118 \realname{} tasks, sampling independent trajectories and using the same model to select among their final laws without evaluator feedback (Appendix~\ref{app:best_of_n_protocol}).
From \(N=5\) to \(N=20\), the metric-specific oracle scores rise by 4.5 points in \(S_N\) and 5.3 points in \(S_V\), while self-selected performance improves by less than 2 points on either metric (Figure~\ref{fig:repeated_sampling}).
Within the five-candidate groups presented to the selector, 60\% of selected winners are Pareto-dominated by another candidate, which scores no lower on either metric and strictly higher on at least one.
These missed improvements cannot be explained solely by a trade-off between predictive fit and scientific validity.
Together, the two analyses show that stronger available candidates do not consistently translate into better final outputs.

\section{Conclusion and Discussion}
We introduced \benchname{}, a curated collection of scientific task packages grounded in the source literature, linking scientific problems, supporting data, published reference equations, and scientific-validity rubrics.
These packages support two complementary settings: \realname{} fixes existing scientific data to evaluate proposed laws, while \parallelname{} fixes synthesized generating equations to evaluate active structural recovery.
Our evaluation yields three main findings.
Predictive fit does not reliably establish scientific validity.
Memorization helps models recover published forms, but full structural recovery remains difficult.
Within a trajectory and across best-of-\(N\) pools, agents evaluate better laws than they submit, and self-selection recovers little of this headroom.
Together, the curated resource and evaluation protocols support studying how models move beyond familiar formulas while remaining grounded in scientific data and constraints.
Our benchmark focuses on closed-form law discovery with specified variables and scientific context, covering only part of the broader discovery process.
Scientific validity is assessed against constraints drawn from the literature, which may not capture every scientifically relevant requirement.
Future work can build on these task packages to develop agents that design informative experiments, reason beyond memorized laws, and reliably identify scientifically valid hypotheses.

\section{Related Work}

\paragraph{Symbolic Regression.} Symbolic regression recovers compact mathematical expressions from observed input--output pairs. Classical methods directly search the expression space using genetic programming, sparse identification, or other program-search procedures \citep{schmidt2009eureqa, cranmer2023pysr, udrescu2020aifeynman, brunton2016sindy, randall2022bingo, burlacu2020operon}, while neural approaches learn priors over equations from large synthetic corpora \citep{biggio2021nesymres, kamienny2022e2esr, valipour2021symbolicgpt}. These methods primarily optimize symbolic formulas against tabular observations, whereas our setting evaluates whether LLMs can combine data with scientific context to discover valid scientific relationships. We therefore do not position classical symbolic regression as our comparison target.

\paragraph{AI for Scientific Discovery.} Recent systems increasingly use LLMs as scientific agents that propose hypotheses, run code, and inspect evidence, including the AI Scientist, AI Scientist-v2, and Co-Scientist \citep{lu2024aiscientist,yamada2025aiscientistv2,gottweis2026coscientist}. A complementary line improves the agents themselves rather than the science: REVERE distills recurring cross-repository failure modes into reusable heuristics that rewrite the agent's own prompts \citep{gangireddy2026revere}. In equation discovery, LLM-SR iteratively refines programmatic equations \citep{shojaee2025llmsr}, SR-Scientist extends this process to longer-horizon code-driven exploration \citep{xia2026srscientist}, and RESTART transfers successful sub-expressions across related problems \citep{li2026robustequationstructure}. These systems motivate evaluating whether LLMs can use scientific context and iterative experimentation, rather than only recover closed-form equations from synthetic samples.

\paragraph{Scientific Discovery Benchmarks.} AI Feynman, SRBench, SRSD, LLM-SRBench, and EmpiricalBench focus on formula recovery \citep{udrescu2020aifeynman,lacava2021srbench,matsubara2022srsd,shojaee2025llmsrbench,cranmer2023pysr}. DiscoveryBench, ScienceAgentBench, MoSciBench, and Stargazer broaden evaluation to data-driven scientific workflows \citep{majumder2025discoverybench,chen2025scienceagentbench,liu2026moscibench,liu2026stargazer}, while SciGym, PhysGym, Gravity-Bench-v1, and NewtonBench place agents in interactive simulated environments \citep{duan2025scigym,chen2025physgym,koblischke2025gravitybench,zheng2026newtonbench}. These benchmarks are useful anchors for formula recovery and active exploration. \benchname targets a different combination: multi-disciplinary empirical laws, real scientific data, domain-knowledge requirements, and paper-grounded validity checks.

\clearpage
\newpage

\appendix
\startcontents[sections]
\printcontents[sections]{}{1}{\section*{Technical Appendices and Supplementary Material}\setcounter{tocdepth}{2}}

\clearpage
\newpage

\clearpage

\section{Residual-Calibrated \parallelname{} Worlds}
\label{app:residual_calibrated_worlds}

This appendix details the construction of the queryable worlds used by \parallelname{}.
Each world combines a synthesized hidden law with an input-dependent residual process estimated from the corresponding source data.
Rather than impose a task-independent parametric noise model, the construction grounds both the input support and residual variation in an empirical scientific problem.
The residual scale is reduced only when necessary to maintain a minimum signal-to-noise ratio under bounded experimentation.

\begin{figure}[H]
\centering
\includegraphics[width=\linewidth]{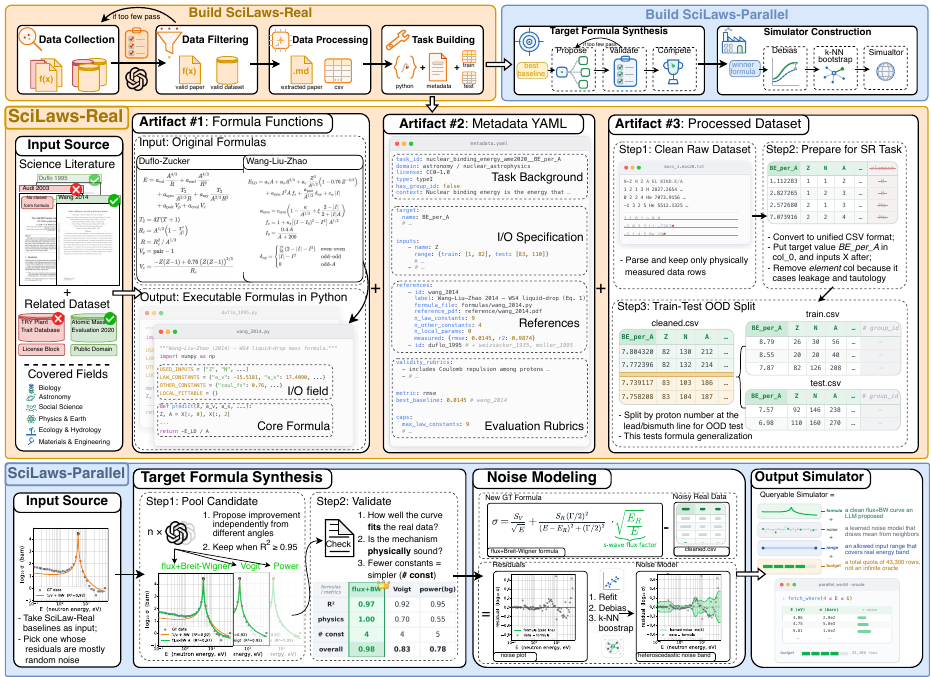}
\caption{Worked example of the \benchname{} construction pipeline, from paper--dataset curation and a fixed-data \realname{} task to hidden-law synthesis and a residual-calibrated \parallelname{} world.}
\label{fig:pipeline_detailed}
\end{figure}

\subsection{Construction Data and Hidden Law}
\label{app:parallel_construction_data}

For task \(t\), let \(\mathcal{D}_t=\{(x_i,y_i,g_i)\}_{i=1}^{n_t}\) denote the corresponding source data, where \(x_i\) contains the candidate scientific inputs, \(y_i\) is the recorded target value, and \(g_i\) is an optional group identifier.
Simulator construction pools all available records from the predefined \realname{} splits because these records calibrate a separate \parallelname{} world rather than evaluate a submitted law.
The solver receives none of these records and begins its \parallelname{} trial without observations.

We first select a published reference whose functional form provides a suitable starting point for synthesis.
For each nondegenerate reference \(f_b\), its coefficients are refitted and its residuals are computed in the task-specific residual space defined below.
Let \(\widehat m_b(x_i)\) be a local \(k\)-nearest-neighbor estimate of the mean residual at \(x_i\).
For each single-group reference, we compute
\begin{equation}
B_b =
\frac{\operatorname{Var}_i[\widehat m_b(x_i)]}
     {\operatorname{Var}_i[r_i^{(b)}]},
\label{eq:systematic_residual_share}
\end{equation}
and select the form with the smallest \(B_b\) among references with nonnegative residual-space \(R^2\).
For multi-group tasks, both quantities are computed within each group and aggregated by their medians, using a fixed sample of at most \(40\) groups during this selection stage.
If no reference has nonnegative residual-space \(R^2\), the same criterion is applied to all nondegenerate references.
This procedure favors a functional form with little systematic residual structure rather than simply the smallest aggregate prediction error.

Multiple agents then propose structural variants of the selected reference under the source problem's scientific context and validity requirements.
The candidate screen checks execution, predictive adequacy, scientific constraints, and integrity of the symbolic representation.
A separate judge selects the hidden law \(F_t\) among admissible variants according to physical plausibility, simplicity, and consistency with the source problem, using predictive fit only as an eligibility criterion.
The selected variant differs structurally from its published starting point and is fixed before solver evaluation.

\subsection{Coefficient Fitting and Residual Space}
\label{app:parallel_residual_space}

We define a task-specific transformation
\begin{equation}
\phi_t(y) =
\begin{cases}
\log_{10}\!\left(\max\{y,\epsilon\}\right) & \text{logarithmic residual space},\\
y & \text{linear residual space}
\end{cases},
\label{eq:residual_transform}
\end{equation}
where \(\epsilon\) is a small positive floor.
Logarithmic residuals are used for \texttt{log\_mae} tasks and for positive, multi-decade targets evaluated with a relative-error metric.
All other tasks use linear residuals.

For single-group tasks, the shared constants of \(F_t\) are refitted by nonlinear least squares in the selected residual space, and the refit is retained only when it does not reduce residual-space \(R^2\).
For multi-group tasks, the declared per-group parameters are estimated separately within each group while shared constants remain common across groups.
Groups with fewer than eight usable records are omitted because they do not support both per-group fitting and residual calibration.
Writing \(\widehat\theta_{t,g}\) for the resulting parameter values, the residual associated with record \(i\) is
\begin{equation}
r_i =
\phi_t(y_i) -
\phi_t\!\left(F_t(x_i;\widehat\theta_{t,g_i})\right).
\label{eq:parallel_residual}
\end{equation}

\subsection{Debiased Local Residual Bootstrap}
\label{app:parallel_residual_bootstrap}

Raw residuals contain both local residual variation and systematic mismatch between the selected law and the source data.
Injecting the systematic component would make the simulator's effective mean response equal to the hidden law plus an unmodeled residual trend.
We therefore estimate the local mean
\begin{equation}
\widehat m_t(x)
=
\frac{1}{|\mathcal{N}_{k}(x)|}
\sum_{j\in\mathcal{N}_{k}(x)} r_j,
\label{eq:local_residual_mean}
\end{equation}
and form debiased residuals \(\widetilde r_i=r_i-\widehat m_t(x_i)\).
Neighbor distances are computed after standardizing each input, with positive inputs spanning more than one decade transformed to logarithmic scale.
The local-mean neighborhood contains approximately \(25\%\) of the available records, subject to sample-size-dependent bounds and a maximum of \(128\) neighbors.

At a query point \(x\), the residual sampler draws uniformly from the debiased residuals in a second local neighborhood:
\begin{equation}
\widetilde R_t(x)
\sim
\operatorname{Unif}\{\widetilde r_j:j\in\mathcal{N}_{k'}(x)\}.
\label{eq:local_residual_bootstrap}
\end{equation}
The sampling neighborhood uses approximately \(15\%\) of the records, clipped to between \(8\) and \(64\) neighbors when the sample size permits.
For smaller samples, the procedure draws from the global residual pool.
This nonparametric bootstrap retains local empirical variation and non-Gaussian residual shape without claiming that the finite records identify the complete measurement process.

\subsection{Noise Calibration and World Definition}
\label{app:parallel_noise_calibration}

The residual magnitude varies widely across scientific problems and can obscure the target law under a bounded query budget.
We therefore allow only downward calibration of the debiased residual scale.
Let \(v_{f,t,g}\) be the variance of \(\phi_t(F_t(X;\widehat\theta_{t,g}))\) over the calibration inputs for group \(g\), and let \(v_{r,t,g}\) be the variance of its debiased residuals.
For target \(\rho=0.9\), the calibration factor is
\begin{equation}
\alpha_{t,g}
=
\min\!\left\{
1,\,
\sqrt{\frac{1-\rho}{\rho}\frac{v_{f,t,g}}{v_{r,t,g}}}
\right\}.
\label{eq:noise_scale}
\end{equation}
For tasks with exactly one input column, \(v_{f,t,g}\) is estimated from points spanning the observed support, using logarithmic sampling when appropriate.
For multivariate inputs, it is estimated by resampling observed input rows so that empirical correlations among inputs are not broken.

The queryable world is defined in residual space by
\begin{equation}
\phi_t(Y_{t,g}(x))
=
\phi_t\!\left(F_t(x;\widehat\theta_{t,g})\right)
+ \alpha_{t,g}\widetilde R_{t,g}(x).
\label{eq:parallel_world}
\end{equation}
Equation~\ref{eq:parallel_world} is additive for linear-space tasks and multiplicative after inversion for logarithmic-space tasks.
A factor \(\alpha_{t,g}=1\) retains the debiased empirical residual scale, whereas \(\alpha_{t,g}<1\) reduces it.
Because the factor is capped at one, the construction never amplifies residual noise and targets residual-space \(R^2\geq 0.9\) rather than forcing every world to have exactly the same \(R^2\).

\subsection{Single-Group and Multi-Group Worlds}
\label{app:parallel_group_structure}

Single-group worlds use one fitted constant set, input support, residual sampler, and calibration factor for the task.
Multi-group worlds use the same hidden functional form in every group but estimate the per-group parameters, input support, residual sampler, and calibration factor separately for each group.
The resulting groups therefore share the recovery target while retaining group-specific parameter values and empirical variation.
Structural evaluation concerns the shared law and its declared pattern of shared constants and per-group parameters, not a model of group identifiers alone.

\subsection{Query and Evaluation Protocol}
\label{app:parallel_query_protocol}

Each trial begins with an empty observation set.
The solver collects measurements by choosing input values and, for multi-group tasks, the group to query.
Inputs outside the calibrated support are clipped to the nearest boundary and reported as such.
Returned measurements accumulate in the solver's analysis environment, while the hidden law, its fitted quantities, and residual sampler remain inaccessible.

All evaluated models receive the same task-specific query limits, which are specified in Appendix~\ref{app:exp_protocol}.

\parallelname{} is scored by structural recovery rather than held-out prediction.
After submission, a code-enabled judge compares the proposed law with \(F_t\) under the ordinal protocol in Appendix~\ref{app:eval_parallel}.
The benchmark does not compute a \parallelname{} numeric-fit or scientific-validity score.

\clearpage
\newpage

\section{Experimental Protocol}
\label{app:exp_details}

\subsection{Base-Model Decoding Settings}
\label{app:exp_config}

The evaluated panel comprises GPT-6, GPT-5.6 Sol, GPT-5.5, GPT-5.4-mini, GPT-5-mini, and GPT-4o-mini~\citep{openai2026models}, together with Claude Opus 4.8~\citep{anthropic2026opus48} and Claude Opus 5.5~\citep{anthropic2026opus55}.
It also includes Gemini 3.5 Flash~\citep{google2026gemini35flash}, DeepSeek-V4 Pro~\citep{deepseekai2026deepseekv4} and DeepSeek-V4.1 Flash~\citep{deepseekai2026deepseekv41flash}, Qwen3.7-Max~\citep{qwenteam2026qwen37max}, Kimi K3~\citep{moonshot2026kimik3}, and GLM-5.2~\citep{zai2026glm52,glmteam2026glm5}.

We hold the agent interface, task inputs, interaction budget, and scoring protocol fixed across models.
Because providers expose different reasoning and decoding controls, we standardize configurations where possible and report the remaining differences explicitly.

Three conventions hold across the whole panel. First, we never
override \texttt{top\_p}, \texttt{top\_k},
\texttt{repetition\_penalty}, \texttt{presence\_penalty}, or
\texttt{frequency\_penalty}: they stay at provider defaults, and
streaming is disabled throughout.

Second, every model with a reasoning head is run at the \emph{medium}
reasoning tier, using whichever knob its provider exposes ---
\texttt{reasoning\_effort=medium} on the GPT-5 series, the equivalent
\texttt{reasoning.effort} field for GLM-5.2, DeepSeek-V4 Pro,
Qwen3.7-Max and Claude Opus 4.8, the corresponding medium thinking
budget for Gemini 3.5 Flash, and the provider's medium reasoning level
for GPT-6, Claude Opus 5.5, Kimi K3, and DeepSeek-V4.1 Flash. Holding the tier fixed is what makes the
rows of Table~\ref{tab:main_results} comparable: the panel is meant to vary
base-model capability, not how much test-time compute each vendor
happens to spend by default. GPT-4o-mini has no reasoning head and is
therefore non-thinking by construction.

Third, decoding parameters follow the reasoning split. Thinking models
receive \emph{no} temperature override --- several providers reject an
explicit value on reasoning endpoints --- and run at the service-side
default ($\approx 1.0$) with a 65{,}536-token completion budget.
GPT-4o-mini, the one non-thinking model, uses
\texttt{temperature=0.4} with an 8{,}192-token cap. The larger budget
is deliberate: on every provider we use, the reasoning trace and the
visible tool block are drawn from the same completion pool, and a
smaller cap truncates \texttt{<python>} bodies mid-expression, which
surfaces as \texttt{finish\_reason=length} rather than as a clean
error.

\subsection{Agent Harness Configuration}
\label{app:exp_harness}

The agent interacts with the task through exactly three XML tools:
\texttt{<python>} for sandboxed analysis, \texttt{<experiment>} for
simulator queries (\parallelname{} only), and
\texttt{<final\_formula>} for submission. The protocol is one tool per
turn; if a response contains several tool blocks, the harness executes
only the first one it sees and ignores the rest. The shared loop runs
for at most \texttt{max\_turns=30} turns per task. The final turn is a
forced-submission turn: the model is told that only
\texttt{<final\_formula>} will be accepted, and if that turn still
fails to produce a parseable submission it is given exactly one retry
before the task is recorded as unsubmitted. Unsubmitted tasks are not
dropped from the denominator: under the strict aggregation of
Appendix~\ref{sec:eval_appendix} they enter every reported mean with a score of
$0$, so a model that runs out of turns is penalised exactly as much as
one that submits an unusable formula.

The \texttt{<python>} sandbox preloads \texttt{train\_df},
\texttt{X\_train}, \texttt{y\_train}, and (on multi-group tasks)
\texttt{group\_ids\_train}, with \texttt{numpy}, \texttt{scipy}, and
\texttt{pandas} already imported. Imports are restricted to a
whitelist (\texttt{numpy}, \texttt{scipy}, \texttt{sklearn},
\texttt{pandas}, \texttt{math}, \texttt{statistics},
\texttt{itertools}, \texttt{functools}, \texttt{collections},
\texttt{warnings}); filesystem, network, and introspection modules are
blocked, as are direct file reads of the task directory, so the agent
cannot recover the held-out split or the hidden generator by any route
other than the tools. Each \texttt{<python>} call is killed after
$100$ seconds of wall time and the timeout is returned to the model as
an ordinary tool error, leaving it free to retry with cheaper code.
Individual API calls time out after $120$ seconds with a single
retry; a call that still fails surfaces as an error inside the loop
rather than aborting the task, so a task is only lost if it exhausts
its turns without submitting.

\subsection{Setting-Specific Protocol}
\label{app:exp_protocol}

In \realname{} the fixed training split is available to the sandbox
from the first turn, and the prompt additionally states the per-column
input ranges of the held-out evaluation split. These ranges make the
intended evaluation regime explicit without revealing the held-out
outcomes. The task-specific splits include representative interpolation,
input-range extrapolation, temporal holdouts, cross-condition
generalization, and, for multi-group tasks, transfer to unseen groups
(Section~\ref{sec:real_setting}).

In \parallelname{} no observations are preloaded. The agent starts
with an empty \texttt{train\_df} and must populate it through
\texttt{<experiment>}, and at least one experiment is required before
\texttt{<final\_formula>} is accepted. Queries are capped along three
axes: at most $10$ \texttt{<experiment>} calls per task, at most
$\max(10, \lfloor 0.1\,n_{\mathrm{train}}\rfloor)$ distinct input
points per call, and at most $3$ repeated samples per point, where
$n_{\mathrm{train}}$ is the task's original training-set size. The
binding constraint is the number of \emph{distinct} design points: ten
calls of $0.1\,n_{\mathrm{train}}$ points each give the agent a design
budget of about $n_{\mathrm{train}}$ points, matching the size of the
\realname{} split it would otherwise have been handed. The extra factor
of three buys only repeated measurements at an already-chosen point,
i.e.\ noise averaging rather than coverage. Structure recovery therefore
has to come from \emph{where} the agent probes, not from how much data it
can accumulate. Each returned batch is
summarised back to the model as a preview of $10$ rows plus batch
statistics, with the full batch left in \texttt{train\_df} for
inspection from the sandbox.

For evaluation we use GPT-5.4-mini as the judge model for the validity
and structure scores, running as a code-enabled judge so that rubric
items requiring probes of the submitted formula can be checked by
execution rather than by inspection alone; the judge prompts and the
verdict vocabulary are given in Appendix~\ref{sec:eval_appendix}.

\subsection{Inference-Time Search and Selection}
\label{app:best_of_n_protocol}

We evaluate inference-time search on all \(118\) \realname{} tasks using
GPT-5.4-mini for both candidate generation and selection. For each task,
we sample \(20\) candidate trajectories under the same task interface and
per-trajectory budget. The \(N=1\) point is the model's standard
single-trajectory result from the main evaluation rather than a member of
the 20-candidate pool. Malformed or non-executing submissions receive zero
under the corresponding official evaluator.

\paragraph{Self-selection.}
The 20 candidates are partitioned into four disjoint groups of five. For
each group, the same model selects one winner after receiving the task
description, full training set, and five candidate programs. The four group
winners define the reported \(N=5\) result, which averages their scores for
each task. They then enter a second selection round, whose winner defines
\(N=20\). The selector receives neither the held-out test set nor scores or
feedback from the official evaluators. All official scores are computed
post hoc.

\paragraph{Metric-specific oracles.}
For a candidate-level quantity \(q\), the \(N=5\) oracle is the exact
expected maximum of \(q\) over a uniformly sampled five-candidate subset of
the 20-candidate pool. The \(N=20\) oracle is the maximum over the full pool.
We apply this definition separately to \(S_N\), \(S_V\), and the
joint-balance score below.

For a task-specific 20-candidate pool \(P_t\), we normalize
\(m\in\{S_N,S_V\}\) as
\[
\widetilde m_t(c)=
\frac{m(c)-\min_{c'\in P_t}m(c')}
{\max_{c'\in P_t}m(c')-\min_{c'\in P_t}m(c')}.
\]
If the denominator is zero, all candidates receive normalized score one.
The separate \(N=1\) baseline is normalized against the same pool range and
clipped to \([0,1]\). Joint balance is
\[
J_t(c)=\min\{\widetilde S_{N,t}(c),\widetilde S_{V,t}(c)\},
\]
and the joint oracle maximizes \(J_t\) under the oracle construction above.

Candidate \(c'\) Pareto-dominates \(c\) when
\(S_N(c')\geq S_N(c)\) and \(S_V(c')\geq S_V(c)\), with at least one strict
inequality. Tied or duplicate candidates therefore do not dominate one
another. Confidence intervals in Figure~\ref{fig:repeated_sampling} use
\(50{,}000\) task-bootstrap replicates.
\subsection{Paired Model Comparison}
\label{app:paired_model_comparison}

To assess GPT-6's advantage over GPT-5.5, we bootstrap their paired task-level scores under the strict scoring convention of Table~\ref{tab:main_results}.
Each of \(20{,}000\) bootstrap replicates samples \(118\) tasks with replacement, retaining both models' scores for each sampled task and giving tasks equal weight.
Using random seed \(20260926\), we obtain percentile \(95\%\) confidence intervals for the mean paired differences.
GPT-6 exceeds GPT-5.5 by \(1.22\) percentage points in \(S_N\) (\(95\%\) CI: \([-1.47, 3.96]\)), \(1.07\) points in \(S_V\) (\([-2.86, 4.90]\)), and \(14.19\) points in \(S_S\) (\([8.90, 19.70]\)).
These intervals quantify variation across resampled tasks conditional on the observed runs, rather than variation across independent agent runs.

\subsection{Scoring Intermediate Candidate Laws}
\label{app:replay_protocol}

For Figure~\ref{fig:discovery_process}(b), each task's candidate pool includes its final submission, so the best-candidate score is the maximum of the intermediate-candidate scores and the final score.
Tasks without a scored intermediate candidate are excluded from this comparison rather than assigned a zero selection gap.

We recover intermediate candidate laws from saved \realname{} single-group trajectories and evaluate those that yield full-length, finite predictions on the held-out split.
We analyze the main-evaluation trajectory of every model on all 66 single-group \realname{} tasks.
Each \texttt{<python>} block that calls a fitting routine is executed twice with the harness sandbox.
In the first run, on the training split, every call to a fitting routine (\texttt{least\_squares}, \texttt{curve\_fit}, \texttt{minimize}, \texttt{lstsq}, \texttt{polyfit} and variants) is recorded in call order, and the printed output is compared with the trajectory's original tool output to validate the procedure.
In the second run the sandbox holds the test split instead, fitting calls return the recorded train-fitted results rather than refitting, and the target vector is instrumented so that every array the agent's own code subtracts from it, i.e.\ the residual of a candidate, is captured as that candidate's test-split prediction; monotone transforms the agent applied to the target (logarithms, square roots, affine rescaling) are inverted before scoring.
Each captured full-length, finite prediction is scored with the task's official metric and reference anchor, giving the \(S_N\) the same law would receive as a submission.
Candidates fitted on a subset of the training data (cross-validation folds) and constant predictors are excluded.
The median agreement between re-executed and original printed numbers is 1.00, and the submitted law of every trajectory receives its stored score.
Multi-group tasks and \parallelname{} are not analyzed this way, because their scores depend on per-group calibration and on the structure judge respectively.

\clearpage
\newpage

\section{Evaluation Protocol and Verbatim Judge Prompts}
\label{sec:eval_appendix}

This appendix specifies the setting-specific evaluation protocols and reproduces the complete prompts used by the two code-enabled judges.
\realname{} reports \texttt{numeric\_score} and \texttt{validity\_score} separately, whereas \parallelname{} reports \texttt{structure\_score} for recovery of the hidden simulator mechanism.

\subsection{\realname{} Reference-Relative Numeric Scoring}
\label{app:eval_numeric}

The numeric channel asks whether a submitted expression predicts held-out data better than the strongest published formula collected for the task.
Each task specifies one primary raw metric from \{\texttt{rmse}, \texttt{smape}, \texttt{log\_mae}\} (lower is better) or \texttt{r2} (higher is better).
The perfect value is \(0\) for the first three metrics and \(1\) for \texttt{r2}.

Let \(\mathcal{B}_t\) denote the set of published formulas proposed by domain researchers and collected for task \(t\).
We evaluate every formula under the same split and fitting protocol as submissions and select the strongest reference:
\[
b_t^\star =
\begin{cases}
\arg\min_{b \in \mathcal{B}_t} m(b), & \text{lower-is-better metric},\\
\arg\max_{b \in \mathcal{B}_t} m(b), & \text{higher-is-better metric}.
\end{cases}
\]
Its raw metric value \(m(b_t^\star)\) defines \(m_{\mathrm{ref}}\).
A submission with raw metric \(m_{\mathrm{sub}}\) receives
\[
S_N(t) = \operatorname{clip}\!\left(1-\frac{1}{2}
\frac{m_{\mathrm{sub}}}{m_{\mathrm{ref}}},0,1\right)
\]
for lower-is-better metrics, and
\[
S_N(t) = \operatorname{clip}\!\left(0.5+\frac{1}{2}
\frac{m_{\mathrm{sub}}-m_{\mathrm{ref}}}
{1-m_{\mathrm{ref}}},0,1\right)
\]
for higher-is-better metrics.
The clipping operator bounds \(S_N(t)\) to \([0,1]\).
The strongest collected formula scores \(0.5\), and a perfect predictor scores \(1.0\).

\paragraph{Single-group tasks.}
The submitted module declares the columns it uses and defines a single
\texttt{predict} function. The evaluator builds the held-out input
matrix in that declared column order and computes the task's raw metric
on the held-out test split. Contract violations, failed imports,
runtime errors, shape mismatches, and non-finite predictions receive a
zero numeric score.

\paragraph{Multi-group tasks.}
Multi-group tasks test whether a functional form transfers across groups
while allowing a small number of per-group parameters to vary. The
submission declares \texttt{LOCAL\_FITTABLE} parameters and supplies a
\texttt{fit} routine. For each test group, the evaluator fits those
per-group parameters on the group's fit window and then evaluates
\texttt{predict} on the held-out window from the same group. The
group scores are averaged with equal weight. Failed groups score
\(0\). If the best reference is already effectively perfect on a
group, that group is excluded because it no longer discriminates
between submissions.

\subsection{\realname{} Validity Judge}
\label{app:eval_validity}

The validity channel asks whether a formula behaves like a plausible
scientific law, rather than merely fitting the held-out observations.
Each task contains frozen validity rubrics derived from the source
problem. Rubrics include behavioral checks, such as monotonicity,
finite outputs, non-negativity, bounds, limits, and separability, and
structural checks, such as required variable dependencies or the
presence of a physically meaningful term. Coefficient accuracy is
intentionally left to \texttt{numeric\_score}; validity focuses on
functional behavior and scientific form.

We use a code-enabled judge because many rubric items require evaluating the submitted formula on controlled probes.
For each task, the judge receives the metadata, observations, rubric list, and submitted formula, and returns rubric-level verdicts in JSON.
The complete evaluator prompt is reproduced below; it is quoted verbatim and
uses \emph{cluster} for what the main text calls a \emph{group}.

\begin{promptbox}
\begin{promptlisting}
You are a VALIDITY JUDGE for SciLaws-Real.

Read only the staged task directories listed below. Write only JSON result
files under the output directory. Do not edit existing repository files.

For EACH staged task:
1. Read validity_rubrics.json and score the rubric list.
2. Read metadata.yaml for target, inputs, and task type.
3. Read and execute submission.py in an isolated namespace. Extract predict,
   USED_INPUTS, LAW_CONSTANTS, and, for multi-group tasks, fit and LOCAL_FITTABLE. If
   import or contract execution fails, write validity_score=null with an
   error string.
4. Build X from the declared USED_INPUTS.
   - Single-group: evaluate the held-out test data.
   - Multi-group: group the fit split by group_id, call fit on representative
     clusters, then call predict with the fitted local parameters.
5. Build deterministic domain grids over each used input's observed min/max
   range while holding other inputs at their median.
6. Score each rubric as Y or N. Prefer computed evidence:
   - behavioral: finite-difference monotonicity, min/max range checks, sign,
     non-negativity, bounds, limits, mixed differences for separability.
   - structural: numeric probes where possible, otherwise source inspection.
     Functionally equivalent terms count; coefficient accuracy belongs to
     numeric_score, not validity_score.
   - constant-discipline / no-cap-evasion: use source inspection, metadata
     caps, and final submitted code. Mark N for fitted/data-derived lookup
     tables, profiles, train/test aggregates, large literal arrays, or obvious
     attempts to move degrees of freedom outside the stated caps.

Write exactly one JSON per task:
{
  "task": "<stage_id>",
  "n_satisfied": <int or null>,
  "n_total": <int or null>,
  "validity_score": <number or null>,
  "error": <string or null>,
  "rubrics": [
    {"i": 1, "verdict": "Y|N", "kind": "behavioral|structural",
     "evidence": "one-line computed or source evidence"}
  ]
}
\end{promptlisting}
\end{promptbox}

For a task with \(N\) applicable rubric items and \(M\) satisfied
items, the score is \(M/N\). The aggregate \texttt{mean\_scored}
averages finite task scores. The stricter aggregate
\texttt{strict\_mean} averages all staged tasks and treats missing,
null, or errored scores as \(0\).

\subsection{\realname{} Anti-Hacking Rubric}
\label{app:eval_antihacking}

Closed-form discovery benchmarks are vulnerable to submissions that
hide memorization or high-dimensional fitting capacity inside code
while still presenting a superficially simple \texttt{predict}
function. The evaluator therefore appends a global anti-hacking
rubric to every task. The rubric does not penalize legitimate fitted
constants; instead, it asks whether the submission circumvents the
declared law contract.

The judge marks this rubric as failed when a submission stores many
extra degrees of freedom outside the declared interface, for example:
large literal arrays, lookup tables keyed by input values, copied
training or test aggregates, per-group profiles, excessive
\texttt{LAW\_CONSTANTS}, too many \texttt{LOCAL\_FITTABLE}
parameters, oversized initialization lists, or other code paths that
approximate the dataset rather than express a law. This catches the
failure mode in which a model writes many parameters to force a good
numeric fit instead of discovering a compact functional relationship.
During aggregation, a failed anti-hacking verdict hard-gates the final
\texttt{validity\_score} for that task to \(0\).

\subsection{Fit--Validity Concordance}
\label{app:fit_validity_concordance}

For each task, we form every unordered pair of executable model submissions
with unequal numeric-fit scores. We orient each pair so that submission
\(i\) has the higher numeric fit,
\(\Delta S_N=S_N^{(i)}-S_N^{(j)}>0\), and retain it when
\(\Delta S_N\geq\tau\). Its tie-aware concordance contribution is
\[
h_{ij} =
\begin{cases}
1, & S_V^{(i)}>S_V^{(j)},\\
\frac{1}{2}, & S_V^{(i)}=S_V^{(j)},\\
0, & S_V^{(i)}<S_V^{(j)}.
\end{cases}
\]
Here \(S_V\) is the official validity score after applying the anti-hacking
gate. We report the mean of \(h_{ij}\) over all retained task--model pairs.
At \(\tau=0\), this includes \(9{,}279\) pairs. The \(50\%\) reference
corresponds to no directional association between the two rankings, with
validity ties divided equally between agreement and disagreement. Confidence
intervals are obtained from \(20{,}000\) bootstrap samples of tasks, retaining
all eligible model pairs within each sampled task.

\subsection{\parallelname{} Structure-Recovery Judge}
\label{app:eval_parallel}

\parallelname{} is scored against the hidden simulator mechanism, not
against a held-out prediction metric. During a trial, the solver sees
only public task metadata and may collect observations by querying the
simulator. After the final formula is submitted, the evaluator stages
three files for each task: \texttt{metadata.yaml}, the hidden
\texttt{formula.py} defining \(F_t\), and the solver's
\texttt{submission.py}. A code-enabled judge then assigns an ordinal
\texttt{structure\_score}.

The allowed scores are:
\[
\{0.00, 0.25, 0.50, 0.75, 1.00\}.
\]
The levels correspond to: unrelated or invalid; relevant variables or
trend only; published form recovered; published form plus
most important added terms recovered; and full hidden-structure
recovery up to algebraic equivalence with consistent coefficient signs and scales. The judge may use source
inspection and small diagnostic probes over the metadata input ranges,
but it is instructed not to compute prediction scores, validity
scores, test RMSE, or leaderboard-style performance. This prevents
dense simulator probing from being rewarded unless it leads to the
right governing law.

The complete evaluator prompt is reproduced below.

\begin{promptbox}
\begin{promptlisting}
You are a PARALLEL STRUCTURE JUDGE for SciLaws-Parallel.

Read only the staged task directories listed below. Write only JSON result
files under the output directory. Do not edit existing repository files.

Goal:
Score whether the submitted formula recovers the hidden simulator GT mechanism
structure. This is NOT a prediction-score task.

Allowed structure_score values are exactly:
- 0.00: unrelated / invalid
- 0.25: relevant variables or trend only
- 0.50: published base form recovered
- 0.75: published base form + most added terms recovered
- 1.00: full hidden structure up to algebraic equivalence, with consistent coefficient signs/scales
        consistent

For EACH staged task:
1. Read task_dir/metadata.yaml for target and input names.
2. Read task_dir/formula.py as the hidden simulator GT. Treat this as the
   reference mechanism, not as public solver context.
3. Read task_dir/submission.py as the solver answer.
4. Compare GT and submission by source inspection and, if useful, small
   diagnostic Python probes over metadata input ranges. Do not compute or report
   prediction scores, validity scores, test RMSE, or leaderboard-style
   performance.
5. Count algebraic equivalence as correct: renaming helper variables, factoring
   terms, log-base changes with transformed coefficients, and equivalent
   reparameterizations should not be penalized.
6. Penalize missing mechanism terms even if predictions are close, including
   task-specific correction terms, interaction terms, saturation terms,
   piecewise regimes, offsets, exponents, and nested transforms.
7. Coefficients should affect only the jump from 0.75 to 1.00 unless the wrong
   sign or order of magnitude changes the mechanism.

Write exactly one JSON per task:
{
  "task": "<stage_id>",
  "structure_score": 0.0 | 0.25 | 0.5 | 0.75 | 1.0,
  "level": "unrelated_or_invalid | variables_or_trend_only | main_structure |
            main_plus_extra_terms | full_gt_structure",
  "error": <string or null>,
  "gt_summary": "one-line GT mechanism summary",
  "submission_summary": "one-line submitted mechanism summary",
  "matched": ["short evidence bullets"],
  "missed": ["short missed-structure bullets"],
  "coefficient_assessment": "short note on signs/scales/equivalence"
}
\end{promptlisting}
\end{promptbox}

Contract failures, missing \texttt{predict} functions, wrong target
shapes, invalid judge outputs, or failed imports receive
\texttt{structure\_score}=0. The aggregate report includes the mean
over finite judge outputs and a strict mean that treats missing,
failed, or errored judgments as zero.

\section{Human Validation of the LLM Judges}
\label{app:human_alignment}

Three of the scores in this paper are produced by an LLM judge rather than
computed in closed form: the \realname{} validity score \(S_V\) and the
\parallelname{} structure score \(S_S\) in Table~\ref{tab:main_results} are assigned
by \texttt{gpt-5.4-mini}, and the cold-recall memorization audit
(Section~\ref{sec:memorization}) uses \texttt{gpt-4.1} to decide, for each data-free
model output, whether it matches the published reference form. The task-level
cold-recall verdict is not itself an LLM decision: it follows deterministically
from five such judgments by a \(\ge\!3/5\) majority rule (Appendix~\ref{app:memorization}).
Because these judges stand in for human scientific judgment, we validate all
three against people. Five domain experts independently re-labeled a stratified
sample of each judge's own decisions, blind to the judge's label and to the model
identity, and we compare the judge--human agreement against the agreement among
the experts themselves. On all three dimensions the judge--human agreement falls
within the range of pairwise human--human agreement (Table~\ref{tab:human_align}), which supports using the judges to score the
benchmark at scale.

\subsection{Sampling design}
\label{app:ha_sampling}

We use a two-stage stratified sampling design. First, we draw the same number of
items from every model --- \(15\) from each of nine evaluated models, for
\(135\) items per score --- so the judge is checked equally on strong and weak
models, rather than mostly on whichever model happens to produce many borderline
answers. Second, within each model we spread those items evenly across the
judge's own output levels: the two memorization match classes, the four \(S_V\)
score bins, and the ordinal \(S_S\) levels. The lowest structure level, \(S_S=0\)
(unrelated/invalid), is folded into the adjacent \(\le\!0.25\) stratum, since the
judge assigns it on only about two benchmark tasks. Spreading a fixed budget evenly across
strata, rather than in proportion to how common each stratum is, is the standard
design when every stratum should be estimated with comparable
precision~\citep{cochran1977sampling}; as a side effect it keeps the label
classes roughly balanced, so the agreement statistics below are not dominated by
a single common class. Within each stratum, items are spread further across the
six scientific domains and the single- and multi-group task types. The draw is
made by a fixed, seeded script, and the item sheets given to the experts carry no
model identity and no judge label. Because the sample is stratified by design
rather than drawn in proportion to natural prevalence, the statistics below
describe the judges' reliability on this validation sample --- which deliberately
covers both common and borderline cases --- rather than their average agreement
over the raw benchmark output.

\subsection{Experts and protocol}
\label{app:ha_protocol}

\begin{figure*}[!h]
\centering
\includegraphics[width=\linewidth]{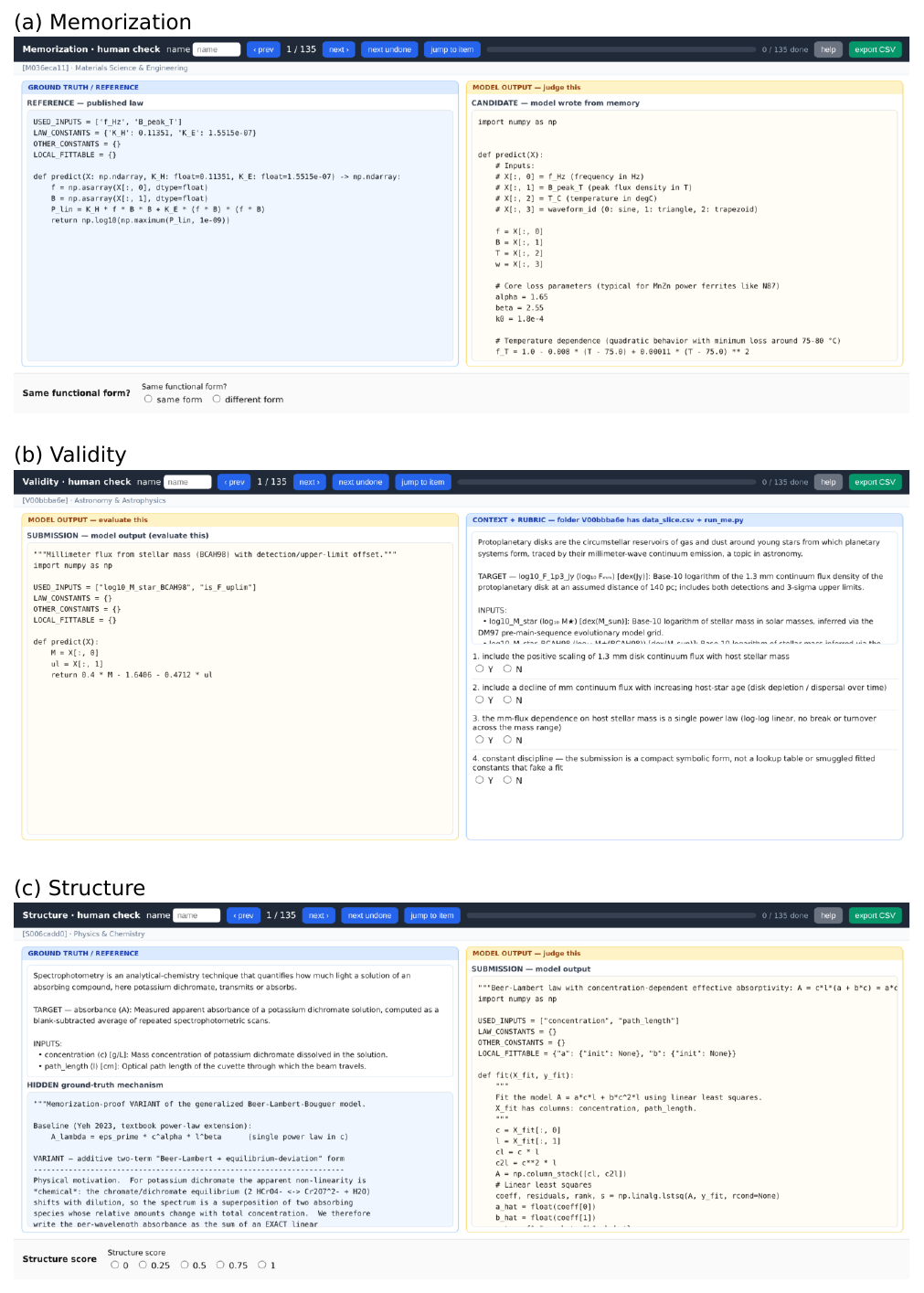}
\caption{\textbf{The offline expert annotation interface}, showing one
representative item for \textbf{(a)} memorization, \textbf{(b)} validity \(S_V\),
and \textbf{(c)} structure \(S_S\).}
\label{fig:human_align_ui}
\end{figure*}

Five researchers with scientific-domain training independently labeled all
\(135\) items of each score in the same offline interface
(Figure~\ref{fig:human_align_ui}). Each item presents the same task artifacts the judge
scored and asks for the same decision, with the judge's label and the model
identity withheld and the item order shuffled. For memorization, the published reference law is shown beside
the model's from-memory formula, and the annotator marks whether the two share
the same functional form. For validity, the model's submission is shown beside
the task's rubric checklist, and each rubric item is marked satisfied or not; the
validity score is the fraction satisfied. For structure, the hidden ground-truth
mechanism is shown beside the model's submission and scored on the five ordinal
levels \(\{0, 0.25, 0.5, 0.75, 1\}\). We take the majority vote of the five
experts as the human label (the median if there is no majority) and compare it against the held-back judge label; as a
reference for how much agreement to expect, we also measure how well the experts
agree among themselves.

\subsection{Metrics}
\label{app:ha_metrics}

We report two quantities.
The first is the raw \emph{agreement} rate --- the fraction of items on which the judge's label matches the aggregated human label --- which is the quantity reported for this kind of validation~\citep{zheng2023judging}.
The second is \emph{Cohen's \(\kappa\)},
which corrects agreement for what chance alone
would produce~\citep{cohen1960coefficient}. The structure score is ordinal, so for
it we use the quadratic-weighted \(\kappa\)~\citep{cohen1968weighted}, which
penalizes larger ordinal disagreements more heavily than adjacent-level ones. We
read each value against \emph{human--human agreement} --- the same statistic
computed between experts, i.e.\ the ten pairwise \(\kappa\) values among the five
annotators --- following the standard practice that a judge is reliable when its
agreement with people is comparable to the agreement among the people
themselves~\citep{zheng2023judging}. Each judge \(\kappa\) is reported with its
\(95\%\) bootstrap CI (Table~\ref{tab:human_align}), which we read against the
human--human range rather than over-interpreting small differences between the
point estimates.

\subsection{Results}
\label{app:ha_results}

Table~\ref{tab:human_align} reports the outcome. On
all three dimensions the judge--human agreement lies within the range of pairwise
human--human agreement: memorization \(\kappa = 0.77\) (human--human
\(0.62\)--\(0.88\)), validity \(\kappa = 0.82\) (\(0.82\)--\(0.91\)), and structure
quadratic-weighted \(\kappa = 0.84\) (\(0.74\)--\(0.90\)). Exact agreement is
\(89\%\) for memorization, \(92\%\) for validity, and \(66.7\%\) for the five-level
structure score. The judge's disagreement with the human consensus is therefore
comparable in size to the disagreement among the human annotators, and it is not
driven by a single lenient rater: each of the five experts individually agrees
with the judge at \(\kappa = 0.66\)--\(0.88\).

\begin{figure}[H]
\centering
\includegraphics[width=0.75\linewidth]{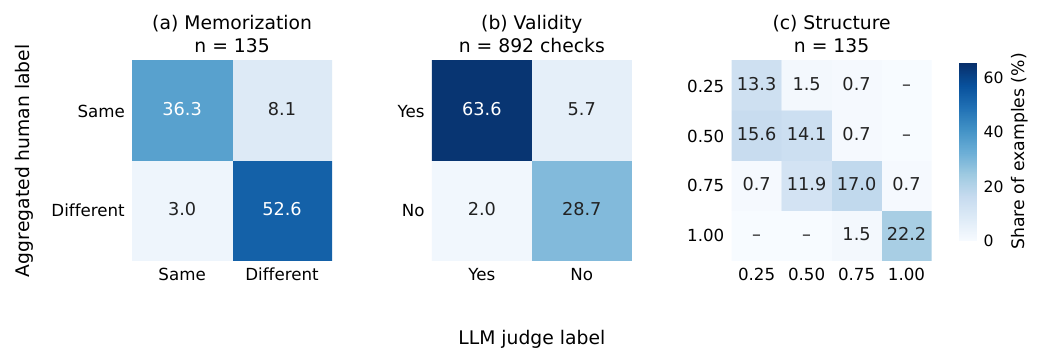}
\caption{\textbf{Confusion between aggregated human (rows) and LLM-judge (columns) labels.}
Cells are percentages of items; outlined cells are exact matches and dashes denote zero.}
\label{fig:human_align_confusion}
\end{figure}

Figure~\ref{fig:human_align_confusion} shows where the residual disagreement lies.
For all three scores the mass is on the diagonal, where the judge's label matches the aggregated human label.
The judge's structure score differs from the aggregated human score by at most one level on \(98.5\%\) of the \(135\) annotated examples, with only two differing by more than one level.
Exact match therefore understates
agreement on structure, since it treats an adjacent-level miss the same as a
distant one; the quadratic-weighted \(\kappa\) is the appropriate summary and is
the value reported in Table~\ref{tab:human_align}. Taken together, the judge--human
agreement is comparable to the human--human agreement on every dimension, which
supports using the judges to score the benchmark at scale.

\begin{table}[h]
\centering
\caption{\textbf{Human validation of the three LLM judges against labels aggregated from five experts.}
\(N\) counts items for
Memorization and Structure and rubric checks for Validity. Judge \(\kappa\) is
shown with its \(95\%\) bootstrap-CI half-width (\(\pm\); a cluster bootstrap over
the \(135\) submissions for Validity). \(^{*}\)\(98.5\%\) of Structure items agree
within one level. \(^{\dagger}\)Quadratic-weighted \(\kappa\).}
\label{tab:human_align}
\begin{tabular}{@{}llcccc@{}}
\toprule
Score & Judge & \(N\) & \shortstack{Exact\\agreement} & Judge--human \(\kappa\) & Human--human \(\kappa\) \\
\midrule
Memorization      & \texttt{gpt-4.1}      & \(135\) & \(89\%\)              & \(0.77 \pm 0.11\)          & \(0.76\ [0.62, 0.88]\) \\
Validity \(S_V\)  & \texttt{gpt-5.4-mini} & \(892\) & \(92\%\)              & \(0.82 \pm 0.05\)          & \(0.87\ [0.82, 0.91]\) \\
Structure \(S_S\) & \texttt{gpt-5.4-mini} & \(135\) & \(66.7\%^{*}\)        & \(0.84^{\dagger} \pm 0.05\) & \(0.84\ [0.74, 0.90]\) \\
\bottomrule
\end{tabular}
\end{table}

\clearpage
\newpage

\section{Memorization Audit: Protocol and Detailed Results}
\label{app:memorization}

\subsection{Purpose and Decision Unit}
\label{sec:mem_new_unit}

A scientific law discovery benchmark using published research must separate two channels that can
produce the same benchmark score. A model may recall a published
reference formula from pretraining, or it may discover a useful form
from the records shown in the task. The memorization audit measures
the first channel directly, without giving the model any data.

The decision unit is a task-model cell \((\tau, M)\). Each
\realname{} task is collapsed to its best-baseline representative
\(f^\star_\tau\), the published reference that the main benchmark
asks a method to beat. This target choice aligns the audit with the
task-level score \(S_N(t)\) and with the above-reference share (\(S_N>0.52\)): recalling a
weaker sister formula may show background knowledge, but it does not
explain a model exceeding the benchmark anchor.

The latest auditable panel contains 118 \realname{} tasks and the
fourteen main-table models spanning seven vendors, giving 1652 task-model
cells. Of these tasks, 66 are single-group and 52 are multi-group.

\subsection{Cold-Recall Probe}
\label{sec:mem_new_probe}

The probe uses a restricted, data-free subset of the benchmark
information. The model sees the target quantity and the ordered input
variables, including names, symbols, units, descriptions, and typical
ranges. It sees no rows, no citation, and no paper text. It is asked
to output a single Python \texttt{def predict(X)} function. For each
task-model cell we sample five completions, at temperature \(0.8\) where the request sets one and at the provider default otherwise.

The prompt asks the model to use only its own knowledge and to write
the standard closed-form relationship for the quantity. This is
intentionally data-free: a hit means that the model can produce the
reference formula from this restricted task description without access
to task data.

\begin{promptbox}
\begin{promptlisting}
You are an expert in {domain}. Using ONLY your own knowledge --
you are given NO data -- write the standard closed-form
relationship for the quantity below as a Python function.

Quantity to predict:
- {target_name} ({target_symbol}) [{target_unit}] -- {target_desc}

Inputs (X is a 2-D numpy array; column X[:, i] is the i-th input
listed):
{inputs_block}

Write a single Python function `def predict(X):` (use
`import numpy as np`) that returns a 1-D array of {target_symbol}.
Fitted coefficients may be literal numbers or named constants.
If several canonical forms exist, write the most commonly cited one.
Output ONLY one ```python ... ``` block, nothing else.
\end{promptlisting}
\end{promptbox}

\subsection{Structural Judge and Verdict}
\label{sec:mem_new_judge}

Each candidate function is compared with the best-baseline reference
by a calibrated LLM judge (\texttt{gpt-4.1}) at temperature \(0\). The judge decides
structural equivalence of the two Python functions. Free coefficients
may be literal numbers or named constants and are treated as fittable
coefficients. Algebraic rearrangements and coefficient
reparameterizations are allowed. In contrast, exponents, the number
of terms, the variables used, and the functional family are
structural. For example, a linear combination does not match a power
law, and a Gompertz form does not match a logistic form.

A task-model cell is marked as cold-recalled if at least three of the
five cold-recall samples match the reference form:
\begin{equation}
\mathrm{cold\text{-}recalled}(\tau, M) =
\mathbb{1}\left[\#\mathrm{hits}(\tau, M) \ge 3 \right].
\label{eq:mem_new_verdict}
\end{equation}
All other cells are marked as not cold-recalled under this protocol.
This binary rule keeps the headline aligned with the benchmark operating
point and avoids an intermediate label whose meaning would be harder to
use in score decomposition.

\subsection{Main Results}
\label{sec:mem_new_results}

Across 118 tasks and fourteen models, \(31.7\%\) of task-model cells are cold-recalled, while \(68.3\%\) are not.
Observed recall ranges from \(15.3\%\) for GPT-4o-mini to \(44.1\%\) for Claude Opus 5.5.
GLM-5.2 and Qwen3.7-Max each recall \(35.6\%\), while GPT-6 recalls \(33.9\%\).
Even the model with the highest observed recall does not reproduce the reference on more than half of the tasks.

\begin{table}[t]
\centering
\small
\setlength{\tabcolsep}{5pt}
\renewcommand{\arraystretch}{1.05}
\caption{Task-level cold-recall verdicts on the 118-task \realname{}
audit panel, across the fourteen main-table models. Cold-recalled means at
least three structurally equivalent outputs among five data-free
samples.}
\label{tab:mem_new_ladder}
\begin{tabular}{lccc}
\toprule
Model & Cold-recalled (\%) & Not cold-recalled (\%) & Tasks \\
\midrule
GPT-4o-mini         & 15.3 & 84.7 & 118 \\
GPT-5.4-mini        & 29.7 & 70.3 & 118 \\
GPT-5.6 Sol         & 34.7 & 65.3 & 118 \\
DeepSeek-V4.1 Flash & 26.3 & 73.7 & 118 \\
GPT-5-mini          & 31.4 & 68.6 & 118 \\
Qwen3.7-Max         & 35.6 & 64.4 & 118 \\
DeepSeek-V4 Pro     & 31.4 & 68.6 & 118 \\
Claude Opus 4.8     & 33.9 & 66.1 & 118 \\
GLM-5.2             & 35.6 & 64.4 & 118 \\
Gemini 3.5 Flash    & 28.8 & 71.2 & 118 \\
Kimi K3             & 28.8 & 71.2 & 118 \\
Claude Opus 5.5     & 44.1 & 55.9 & 118 \\
GPT-5.5             & 34.7 & 65.3 & 118 \\
GPT-6               & 33.9 & 66.1 & 118 \\
\midrule
All cells                 & 31.7 & 68.3 & 1652 \\
\bottomrule
\end{tabular}
\end{table}

\begin{figure}[t]
\centering
\includegraphics[width=0.74\linewidth]{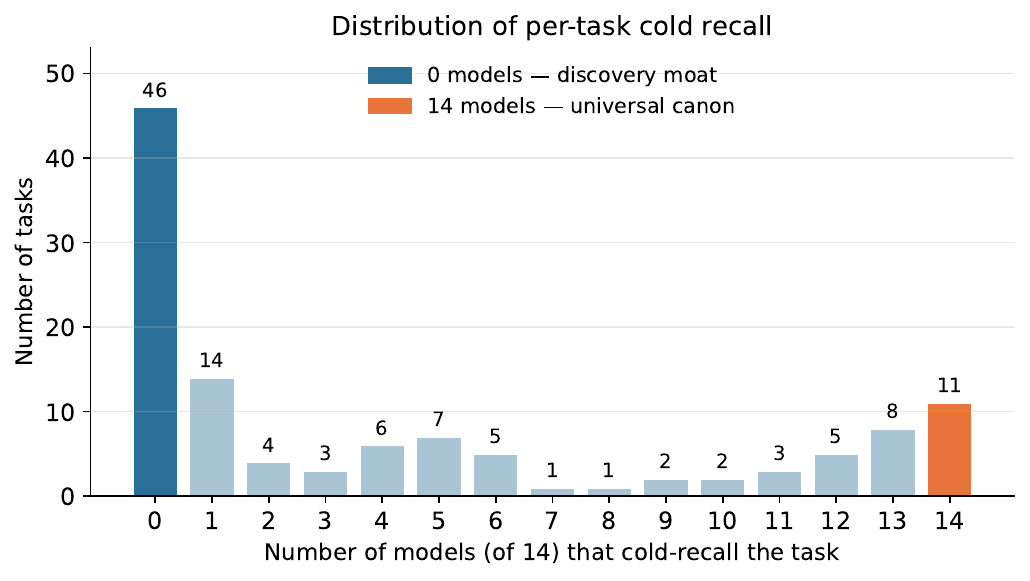}
\caption{Distribution of task-level cold recall across fourteen audited models. The left bar is the 46-task discovery moat; the right bar is the 11-task universal canon.}
\label{fig:mem_new_spectrum}
\end{figure}

The task-level distribution retains a substantial discovery moat (Figure~\ref{fig:mem_new_spectrum}).
No audited model cold-recalls 46 tasks, or \(39.0\%\) of the panel.
At the other end, 11 tasks, or \(9.3\%\), are cold-recalled by all fourteen models.
These universal-canon tasks are AFM spherical indentation, WGS84/Somigliana gravity, Gutenberg--Richter, metabolic scaling, Langmuir CO$_2$ adsorption, Holling functional response, exoenzyme kinetics, Beer--Lambert, at-station hydraulic geometry, trade gravity, and urban scaling.

\subsection{Type and Domain Findings}
\label{sec:mem_new_type_domain}

\begin{figure}[t]
\centering
\begin{minipage}[b]{0.49\linewidth}
\centering
\includegraphics[width=\linewidth]{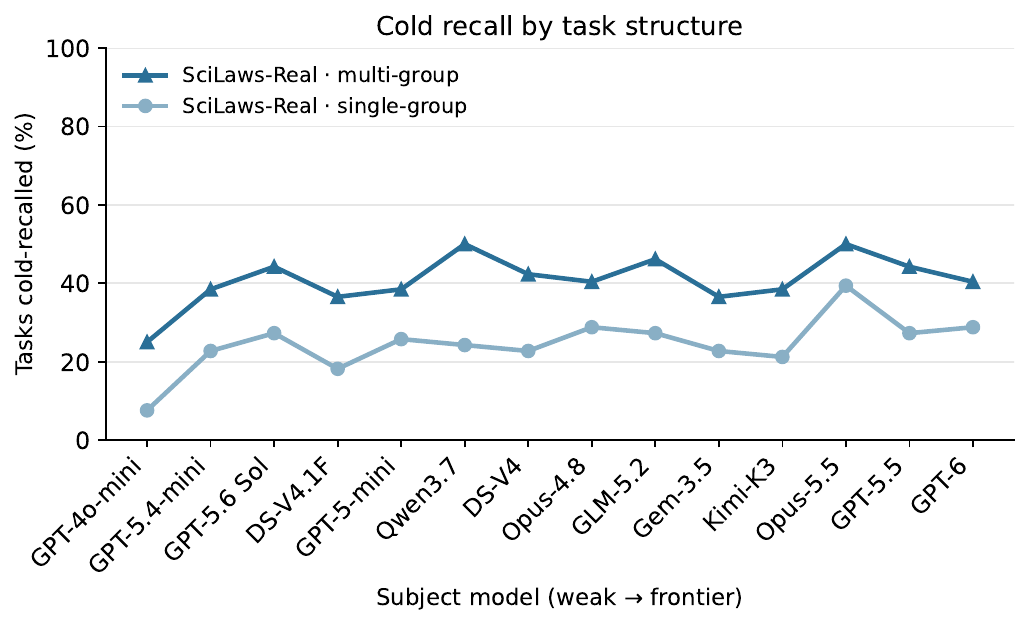}
\caption{Cold recall by task structure. Multi-group tasks are more recallable
than single-group tasks across all fourteen models.}
\label{fig:mem_new_bytype}
\end{minipage}
\hfill
\begin{minipage}[b]{0.49\linewidth}
\centering
\includegraphics[width=\linewidth]{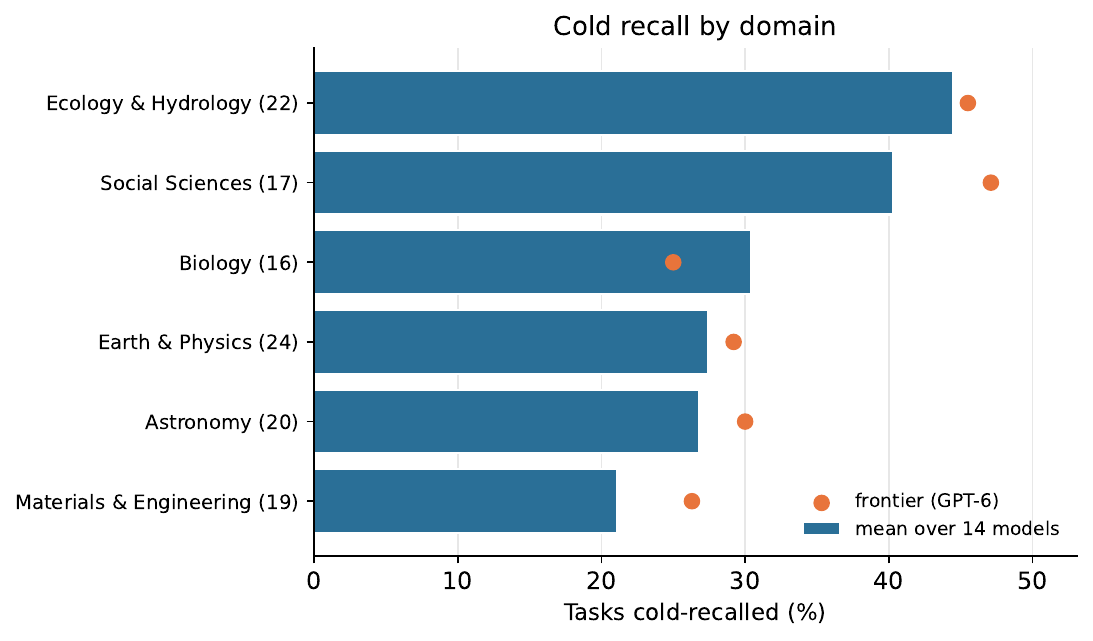}
\caption{Cold recall by six-domain grouping. Bars show the mean over fourteen audited models; dots show GPT-6.}
\label{fig:mem_new_domain}
\end{minipage}
\end{figure}

Multi-group tasks have higher cold recall than single-group tasks: \(40.8\%\)
versus \(24.6\%\) over all task-model cells
(Figure~\ref{fig:mem_new_bytype}). The likely reason is that many multi-group
tasks are organized around named cross-group laws or
invariant-constant forms, which are more likely to appear as
canonical closed forms in training corpora.

The domain split is also uneven (Figure~\ref{fig:mem_new_domain}).
Materials \& Engineering has the lowest mean cold-recall rate, at \(21.1\%\).
Ecology \& Hydrology and Social Sciences have the highest rates at \(44.5\%\) and \(40.3\%\), respectively.
GPT-6 cold-recalls \(45.5\%\) of Ecology \& Hydrology tasks.

\subsection{Target-Choice Robustness}
\label{sec:mem_new_target_robustness}

The headline uses the best-baseline representative as the recall
target. This is the right operating target, but it could undercount
recall if several nearly equivalent published baselines fit the same
task equally well. We therefore ran a relaxed target analysis. Any
baseline with \(R^2 \ge 0.9\) is treated as an additional valid recall
target, and the task is marked as cold-recalled if the model recalls any such
target.

\begin{figure}[t]
\centering
\includegraphics[width=0.92\linewidth]{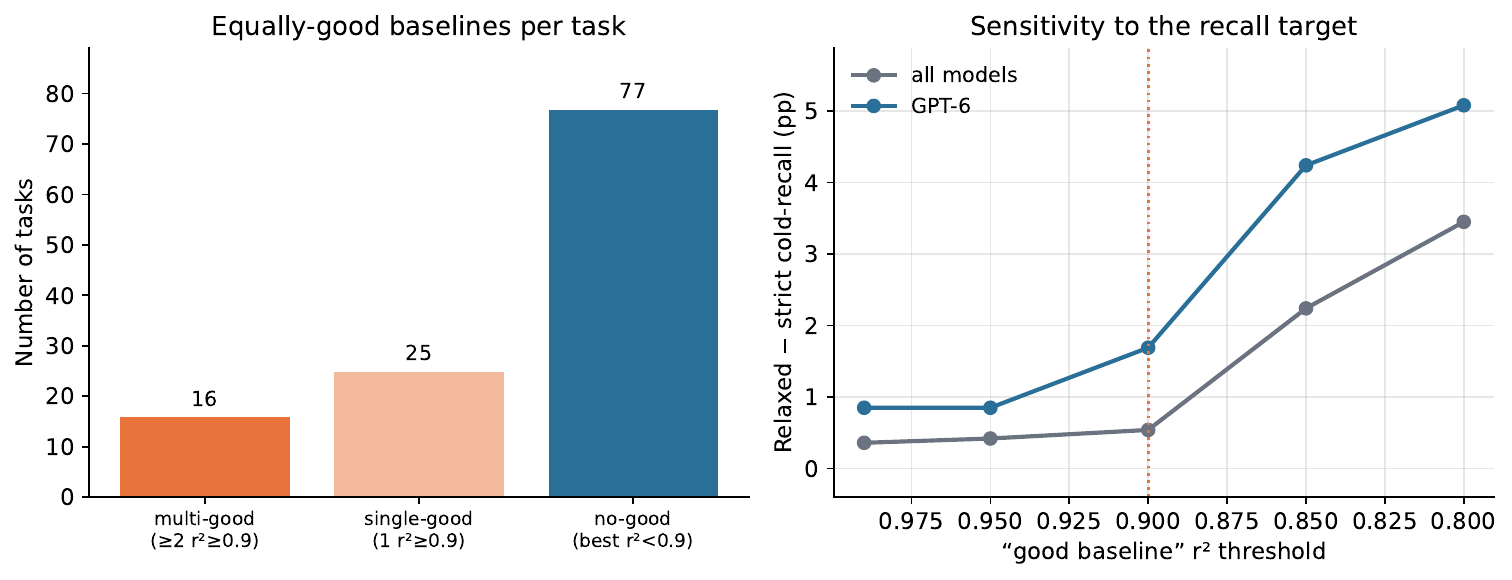}
\caption{Robustness to the recall target: relaxing it from the best baseline to any \(R^2 \ge 0.9\) baseline.}
\label{fig:mem_new_relax}
\end{figure}

Across the fourteen audited models, the relaxed rule changes the recall rate very little
(Figure~\ref{fig:mem_new_relax}). At the \(R^2 \ge 0.9\) threshold, only 16
of 118 tasks have two or more equally good baselines. The all-model
cold-recall rate increases by only \(0.54\) percentage points, and the
GPT-6 rate increases by \(1.7\) points, or two tasks. Only nine task-model
cells flip, across four tasks. This shows that the best-baseline anchor is
not materially hiding recall.

\subsection{AI-Feynman Calibration}
\label{sec:mem_new_feynman}

We audit AI-Feynman~\citep{udrescu2020aifeynman} with the same cold-recall prompt, judge, and five-sample rule using ten OpenAI models evaluated on both corpora: GPT-4o-mini, GPT-4o, GPT-4.1-mini, GPT-4.1, GPT-5-nano, GPT-5-mini, GPT-5, GPT-5.5, GPT-5.6~Sol, and GPT-6.
Five of these (GPT-4o-mini, GPT-5-mini, GPT-5.5, GPT-5.6~Sol, and GPT-6) belong to the fourteen-model main panel; the other five were audited only for this calibration.
Under this controlled comparison, the mean cold-recall rate is \(58.8\%\) for AI-Feynman and \(28.3\%\) for \realname{}; the latter is the mean over these ten models and therefore differs from the \(31.7\%\) of the fourteen-model panel.
GPT-6 cold-recalls \(76.0\%\) of AI-Feynman tasks and \(33.9\%\) of \realname{} tasks, a \(42.1\)-percentage-point gap.

This control is useful because AI-Feynman is intentionally
textbook-like. The audit therefore tags a known highly canonical
benchmark as highly recallable, while assigning much lower recall to
\realname{}. That behavior argues against the probe being too lax or
too strict.

\clearpage
\newpage

\section{Verbatim Solver Prompt Templates}
\label{app:prompts}

This appendix reproduces the complete system and user prompt templates presented to solvers in the \realname{} and \parallelname{} experiments.
Placeholders in braces are filled from task metadata, and bracketed single-group and multi-group clauses indicate the mutually exclusive branch used for each task.
Shared protocol text is repeated so that each template records the complete standalone prompt.
These templates are quoted verbatim as shown to solvers: they use \emph{cluster} for what the main text calls a \emph{group}, and \texttt{LOCAL\_FITTABLE} for the per-group parameters.

\subsection{\realname{} System Prompt}
\label{app:prompt_real_system}

\begin{promptbox}
\begin{promptlisting}
# Role

You are a scientific equation-discovery agent.

## Protocol

- Output EXACTLY ONE tool block per turn, and nothing else -- no surrounding
  prose, no Markdown code fences.
- One block per turn even of the SAME type: do NOT emit two `<python>` blocks
  or `<python>` then `<final_formula>` in one reply. If you want to run several
  analyses, combine them into a SINGLE `<python>` block. If you emit more than
  one block, only one is executed and the rest are discarded.
- Tool results are returned on the next turn -- read them, then take your next
  step.
- Never write `<python_output>` or `<experiment_output>` yourself; those tags
  are reserved for harness feedback after a tool runs.
- Use column names exactly as listed in the task message.

## Workflow

You have a generous turn budget. Use `<python>` to inspect the data and fit
constants, and feel free to iterate before submitting. If a fit looks poor, you
can refine the constants or try a different functional form rather than settle
for your first guess. Submit with `<final_formula>` once you have a form you
are satisfied with.

## Tools

### 1. Run Python: `<python>`

This is how you both inspect the data and fit constants. The entire training
set is preloaded, so there is no separate data-request tool.

<python>
...Python analysis code...
</python>

Preloaded variables:
- `train_df`: pandas DataFrame with all training rows, named columns.
- `X_train`: numeric input matrix with shape `(n_rows, n_inputs)`.
- `y_train`: numeric target vector with shape `(n_rows,)`.
- `input_cols`: input column names; `X_train[:, i]` corresponds to
  `input_cols[i]`.
- `target_col`: target column name; `y_train` is `train_df[target_col]`.
- `group_ids_train`: multi-group tasks only; integer group id for each row.

`numpy`, `scipy`, and `pandas` are available. Only `print(...)` output is
returned. Each `<python>` call starts fresh with the preloaded variables plus
`np`/`scipy`/`pd`; redefine helper functions in each call. Python execution is
limited to 100 seconds. Very long stdout is truncated, so print compact
summaries. Prefer vectorized least squares, `scipy.optimize.curve_fit` /
`least_squares`, or a small targeted search over a few candidates.

### 2. Submit Final Formula: `<final_formula>`

[Single-group contract]

<final_formula>
"""Short description."""
import numpy as np

USED_INPUTS = [...]        # columns predict reads, in X-column order
LAW_CONSTANTS = {}         # leave empty -- bake fitted constants into predict
OTHER_CONSTANTS = {}
LOCAL_FITTABLE = {}

def predict(X):
    ...
    return y_pred
</final_formula>

Notes:
- `<final_formula>` ends the trial.
- `predict(X)` takes ONLY `X`; do not add a `group_id` parameter.
- Keep `LAW_CONSTANTS`, `OTHER_CONSTANTS`, and `LOCAL_FITTABLE` as empty dicts.
- Write fitted constants directly as numeric literals inside `predict`.
- `USED_INPUTS` lists the columns `predict` reads; `X[:, i]` is
  `USED_INPUTS[i]`.
- `predict` must return an ndarray of shape `(N,)`, fully numeric, with no
  fitting at evaluation time.

[Multi-group contract]

This is a multi-group task. The data is split into clusters (`group_id`);
your formula must work on clusters that are not in the observations available
to you. Find one functional form that holds across clusters, where a few
parameters are re-fit per cluster.

<final_formula>
"""Short description of the shared functional form."""
import numpy as np

USED_INPUTS = [...]        # columns predict reads, in X-column order
LAW_CONSTANTS = {}         # universal constants shared by all clusters
OTHER_CONSTANTS = {}
LOCAL_FITTABLE = {"a": {"init": None}, "b": {"init": None}}

def fit(X_fit, y_fit):
    ...
    return {"a": a_hat, "b": b_hat}

def predict(X, a, b):
    ...
    return y_pred
</final_formula>

Notes:
- `<final_formula>` ends the trial.
- `LOCAL_FITTABLE` is a non-empty dict of per-cluster parameter names; keep the
  count small and within the task cap.
- `fit(X_fit, y_fit)` receives one cluster's fit window and returns that
  cluster's parameters; `predict(X, **params)` receives them as keyword
  arguments.
- Neither `fit` nor `predict` may take a `group_id` argument.
- Keep `LAW_CONSTANTS` empty unless a constant is truly universal. Anything
  that varies by cluster must go through `LOCAL_FITTABLE` and `fit()`.
- Use `group_ids_train` in the Python sandbox to check whether the shape
  transfers across training clusters before submitting.
- `predict` must return an ndarray of shape `(N,)`; keep `fit()` fast and robust.
\end{promptlisting}
\end{promptbox}

\subsection{\realname{} User Prompt}
\label{app:prompt_real_user}

\begin{promptbox}
\begin{promptlisting}
{problem_statement}

Target: `{target_col}`{target_suffix}

Scoring: your formula is evaluated on a held-out test set by **{metric}** --
{metric_gloss} -- measured against strong published reference formulas.
Optimise this metric.

Use exactly one XML tool per turn: `<python>` or `<final_formula>`. The full
training set is preloaded in the `<python>` sandbox as `train_df` / `X_train` /
`y_train`; inspect it there (`train_df.describe()`, `train_df.corr()`, slicing,
plots-as-stats). {contract_line}
{caps_block}
{multi_group_block}
Inputs to `predict`:
{input_lines}

Training data: {n_train_rows} rows{train_groups_suffix}{ranges_block}

Budget: at most {max_turns} turns.

You can use `<python>` turns to refine your fit before submitting if it helps.
Aim for a formula whose functional form is physically reasonable and behaves
sensibly when extrapolated beyond the inputs you have seen.

IMPORTANT: emit only ONE XML tool block per turn. If you emit multiple tool
blocks, only the first XML tool block is executed and all later blocks are
ignored. Wait for the tool result before deciding the next call.
\end{promptlisting}
\end{promptbox}

\subsection{\parallelname{} System Prompt}
\label{app:prompt_parallel_system}

\begin{promptbox}
\begin{promptlisting}
# Role

You are a scientific equation-discovery agent.

## Protocol

- Output EXACTLY ONE tool block per turn, and nothing else -- no surrounding
  prose, no Markdown code fences.
- One block per turn even of the SAME type: do NOT emit two `<python>` blocks
  or `<python>` then `<final_formula>` in one reply. If you want to run several
  analyses, combine them into a SINGLE `<python>` block. If you emit more than
  one block, only one is executed and the rest are discarded.
- Tool results are returned on the next turn -- read them, then take your next
  step.
- Never write `<python_output>` or `<experiment_output>` yourself; those tags
  are reserved for harness feedback after a tool runs.
- Use column names exactly as listed in the task message.

## Workflow

You have a generous turn budget, but no preloaded training observations. Start
by probing the simulator with `<experiment>`, then use `<python>` to inspect the
data you collected and fit constants. You may iterate between targeted
experiments and Python analysis before submitting. Submit with
`<final_formula>` once you have a form you are satisfied with.

## Tools

### 1. Probe Simulator: `<experiment>`

<experiment>{"<input_col>": [v1, v2, ...], "n_samples": 3, "seed": 0}</experiment>

The content inside `<experiment>` must be valid JSON with literal arrays and
numbers only. Do not use Python expressions such as `range(...)`, list
comprehensions, variables, `np.linspace(...)`, or comments inside the JSON.
Choose input values deliberately to test low, middle, and high regimes. You
must call `<experiment>` at least once before `<final_formula>`. Respect the
experiment budget caps listed in the task's simulator notes. Oversized requests
are rejected without returning data. Returned rows are appended to `X_train`,
`y_train`, and `train_df` for the next `<python>` call.

### 2. Run Python: `<python>`

This is how you inspect the observations you have collected and fit constants.
The cumulative training log starts empty and is populated only by successful
`<experiment>` calls.

<python>
...Python analysis code...
</python>

Preloaded variables:
- `train_df`: pandas DataFrame with observations returned by successful
  `<experiment>` calls so far.
- `X_train`: numeric input matrix with shape `(n_rows, n_inputs)`.
- `y_train`: numeric target vector with shape `(n_rows,)`.
- `input_cols`: input column names; `X_train[:, i]` corresponds to
  `input_cols[i]`.
- `target_col`: target column name; `y_train` is `train_df[target_col]`.
- `group_ids_train`: multi-group tasks only; integer group id for each row.
- `experiment_log`: simulator tasks only; metadata for each successful
  `<experiment>`.
- `experiment_caps`: simulator tasks only; hard limits on points, samples,
  rows per call, and total simulator rows.

`numpy`, `scipy`, and `pandas` are available. Only `print(...)` output is
returned. Each `<python>` call starts fresh with the preloaded variables plus
`np`/`scipy`/`pd`; redefine helper functions in each call. Python execution is
limited to 100 seconds. Very long stdout is truncated, so print compact
summaries. Prefer vectorized least squares, `scipy.optimize.curve_fit` /
`least_squares`, or a small targeted search over a few candidates.

### 3. Submit Final Formula: `<final_formula>`

[Single-group contract]

<final_formula>
"""Short description."""
import numpy as np

USED_INPUTS = [...]        # columns predict reads, in X-column order
LAW_CONSTANTS = {}         # leave empty -- bake fitted constants into predict
OTHER_CONSTANTS = {}
LOCAL_FITTABLE = {}

def predict(X):
    ...
    return y_pred
</final_formula>

[Multi-group contract]

This is a multi-group task. Find one hidden simulator mechanism structure that
holds across clusters; only the `LOCAL_FITTABLE` parameters may change between
them.

<final_formula>
"""Short description of the shared functional form."""
import numpy as np

USED_INPUTS = [...]        # columns predict reads, in X-column order
LAW_CONSTANTS = {}
OTHER_CONSTANTS = {}
LOCAL_FITTABLE = {"a": {"init": None}, "b": {"init": None}}

def fit(X_fit, y_fit):
    ...
    return {"a": a_hat, "b": b_hat}

def predict(X, a, b):
    ...
    return y_pred
</final_formula>

Final-formula notes:
- `<final_formula>` ends the trial.
- Do not pass `group_id` to `predict` or `fit`.
- Single-group tasks bake fitted constants into `predict`.
- Multi-group tasks declare per-cluster parameters in `LOCAL_FITTABLE`, return
  exactly those keys from `fit()`, and use the same functional form for every
  cluster.
- `predict` must return a finite ndarray of shape `(N,)`.
\end{promptlisting}
\end{promptbox}

\subsection{\parallelname{} User Prompt}
\label{app:prompt_parallel_user}

\begin{promptbox}
\begin{promptlisting}
{problem_statement}

Target: `{target_col}`{target_suffix}

Parallel objective: recover the hidden simulator's mechanism structure from
your own experiments. The primary evaluation is `structure_score`, based on
whether the submitted formula matches the simulator's functional form up to
algebraic equivalence; do not optimize for a held-out prediction metric.

Use exactly one XML tool per turn: `<experiment>`, `<python>`, or
`<final_formula>`. No training observations are preloaded. Use `<experiment>`
to collect observations; successful experiment rows are then available in the
`<python>` sandbox as `train_df` / `X_train` / `y_train`. The `<experiment>`
content must be valid JSON literals only; do not use Python expressions such
as `range(...)`, list comprehensions, or `np.linspace(...)` inside it.
{contract_line}
{caps_block}
{multi_group_block}
Inputs to `predict`:
{input_lines}

Training observations available initially: 0 rows.{ranges_block}
{simulator_notes_block}

Budget: at most {max_turns} turns.

You can use `<python>` turns to refine your fit before submitting if it helps.
Aim for a formula whose functional form is physically reasonable and behaves
sensibly when extrapolated beyond the inputs you have seen.

IMPORTANT: emit only ONE XML tool block per turn. If you emit multiple tool
blocks, only the first XML tool block is executed and all later blocks are
ignored. Wait for the tool result before deciding the next call.
\end{promptlisting}
\end{promptbox}

\section{Case Studies}
\label{app:cases}

This appendix grounds the main findings of Section~\ref{sec:analysis} in concrete task instances from the fourteen evaluated models. Table~\ref{tab:case_overview}
maps each instance to the finding it illustrates and to its trajectory below, where one is given. We then give four complete reference trajectories that span the benchmark's two design
axes---\realname{}/\parallelname{} $\times$ single-/multi-group---so that a reader
unfamiliar with the benchmark can follow the entire evaluation protocol, turn by turn, on real agent logs.

\begin{table}[H]
\centering\footnotesize
\setlength{\tabcolsep}{4pt}\renewcommand{\arraystretch}{1.15}
\caption{Task instances behind the main findings. Type is single-group (I) / multi-group (II); memorization tiers follow Section~\ref{sec:memorization}. The reference is the strongest published baseline, so \(S_N=0.5\) ties it. The selection bottleneck is a pool-level effect (Section~\ref{sec:repeated_sampling}).}
\label{tab:case_overview}
\begin{tabularx}{\linewidth}{@{}>{\raggedright\arraybackslash}p{0.2\linewidth} >{\raggedright\arraybackslash}p{0.12\linewidth} >{\raggedright\arraybackslash}X c@{}}
\toprule
Task (Type, domain) & Model(s) & Observation & Trace \\
\midrule
\multicolumn{4}{@{}l}{\textit{Predictive fit does not reliably imply scientific validity or structural recovery (Section~\ref{sec:numeric_vs_validity})}} \\
Gold neutron-capture resonance (I, nuclear physics) & Qwen3.7-Max & Best fit on the task (\(S_N=0.73\)), but the law adds a second resonance near 47\,eV that the data do not contain; it satisfies 2 of 5 rubric items and fails the anti-hacking check, so \(S_V=0\). & --- \\
Proton form factor \(G_E/G_D\) (I, particle physics) & GPT-5.4-mini & Scientifically valid (\(S_V=1.0\)) but below the reference (\(S_N=0.44\)); 8 of 14 models score \(S_N=0\), 7 of them with \(S_V=1.0\). & \ref{app:case_proton} \\
Keeling curve (I, atmosphere) & GPT-5.5 & Above the reference (\(S_N=0.78\)) while recovering only the trend of the hidden law (\(S_S=0.25\)). & \ref{app:case_keeling} \\
Sellmeier dispersion (II, optics) & GPT-5.5 & Below the reference (\(S_N=0.21\)) while recovering the full hidden law (\(S_S=1.0\)). & \ref{app:case_sellmeier} \\
\midrule
\multicolumn{4}{@{}l}{\textit{Memorization supports published forms, not new structure (Section~\ref{sec:memorization})}} \\
WGS84 normal gravity (I, geodesy; Canon) & GPT-5.5 & Recalls the Somigliana formula and only ties the reference (\(S_N=0.50\)). & --- \\
Chiller efficiency (II, engineering; Moat) & GPT-5.5 & Fits a new form that beats the reference (\(S_N=0.81\)). & --- \\
CO\(_2\) adsorption, T\'oth (II, materials; Moat) & GPT-5.5 & Beats the reference (\(S_N=0.53\); 7 of 14 models do) with a form that transfers to unseen groups. & \ref{app:case_co2} \\
Beer--Lambert absorbance (II, spectroscopy; Canon) & all 14 models & Every model recovers the published form (\(S_S\geq0.5\)); none recovers the full hidden law (\(S_S=1\)). & --- \\
\bottomrule
\end{tabularx}
\end{table}

\noindent\textbf{How to read these traces.} Each task is solved by an agent over a sequence of
\emph{turns}.
\begin{itemize}[leftmargin=1.5em,itemsep=1.5pt,topsep=3pt,parsep=0pt]
\item On its turn the agent emits exactly \textbf{one} block: \texttt{<python>} (run code on the
collected data), \texttt{<experiment>} (query the hidden simulator; \parallelname{} only), or
\texttt{<final\_formula>} (submit and end). The harness then replies with \texttt{<python\_output>} /
\texttt{<experiment\_output>}.
\item The complete system and user prompts are in Appendix~\ref{app:prompts}; each box below shows only the
task-specific part plus every turn.
\end{itemize}

\noindent\textbf{Colour tells you who wrote what.}
\begin{itemize}[leftmargin=1.5em,itemsep=1.5pt,topsep=3pt,parsep=0pt]
\item {\color{traceGray}\textbf{Gray}} --- verbatim: the prompt the model saw, the code it ran
(\emph{including its own \texttt{\#} comments}), and the output it got back.
\item \textbf{Black bold} --- our labels: turn markers and header field names.
\item Black roman --- our own words: the header summary of each task, and the final \textsc{Score}.
\item \textit{Black italic} --- our own margin notes and elisions; an elision, e.g.\
\textit{[\,...\,22 lines\,...]}, always states how many lines we cut and what they were.
\item \textcolor{rolePivot}{\textbf{Red}} --- the few places where the interaction \emph{design} changes
across the four traces: the data source, what is submitted, and how it is scored.
\end{itemize}

\noindent Nothing in gray is paraphrased. Because the traces are verbatim, they use \emph{cluster} for what the main text calls a \emph{group}. The proton trace (Appendix~\ref{app:case_proton}) shows \emph{every turn} ---
with only the two marked cuts --- as a reference for the whole protocol; the other three are faithfully
excerpted, with every cut marked, to keep each to about a page.

\subsection{\textsc{Real} $\times$ single-group: the proton form factor}
\label{app:case_proton}
\begin{trajbox}{\texttt{proton\_em\_form\_factor}, GPT-5.4-mini \quad(all 3 turns shown)}
\begin{trajlisting}
|B|TASK|B|       |N|proton_em_form_factor :  G_E/G_D  as a function of  Q^2   (nuclear form factor)|N|
|B|DATA|B|       |HL|37 rows preloaded as train_df / X_train / y_train|HL| |N|; train Q^2 in [0.007,1.83], test [2.07,5.85] (extrapolates)|N|
|B|SUBMIT|B|     |HL|USED_INPUTS, LAW_CONSTANTS={}, ..., def predict(X) with constants baked in as literals; no fit(), no group_id|HL|
|B|CAPS|B|       |N|<= 8 global constants ; max_local_params=0 ; this task must NOT define fit()|N|
|B|SCORED BY|B|  |HL|S_N -- RMSE on the held-out test set, vs. the strongest published reference|HL|

|B|TURN 1 -- AGENT   <python>|B|   |EL|-- inspect the data, then fit three rational families|EL|
import numpy as np, pandas as pd, scipy.optimize as opt
print("cols:", input_cols, "target:", target_col)
print(train_df.head(10).to_string(index=False))
print("\nDescribe:")
print(train_df.describe().to_string())
x = X_train[:,0]
y = y_train
print("\nRanges:", float(x.min()), float(x.max()), float(y.min()), float(y.max()))
# simple candidate fits
def models(params, x):
    # params: a,b,c,d for (1+a x)/(1+b x+c x^2+d x^3)
    a,b,c,d = params
    return (1+a*x)/(1+b*x+c*x*x+d*x*x*x)
def resid(params):
    return models(params,x)-y
# initial guesses from rough physics
inits = [
    np.array([0.0, 0.0, 0.70, 0.0]),
    np.array([-0.1, 0.0, 0.70, 0.0]),
    np.array([0.0, 0.0, 0.60, 0.10]),
    np.array([-0.2, 0.0, 0.60, 0.10]),
]
best = None
for init in inits:
    try:
        res = opt.least_squares(resid, init, max_nfev=20000)
        rmse = np.sqrt(np.mean(res.fun**2))
        print("fit", init, "->", res.x, "rmse", rmse)
        if best is None or rmse < best[0]:
            best = (rmse, res.x)
    except Exception as e:
        print("fit failed", init, e)
print("best", best)
# try physically motivated dipole-like with denominator-only
def model2(p,x):
    a,b,c = p
    return 1.0/(1+a*x+b*x*x+c*x*x*x)
def resid2(p):
    return model2(p,x)-y
for init in [np.array([0.3,0.0,0.0]), np.array([0.5,0.0,0.0]), np.array([0.2,0.05,0.0]), np.array([0.1,0.1,0.0])]:
    try:
        res=opt.least_squares(resid2, init, max_nfev=20000)
        print("model2", init, "->", res.x, "rmse", np.sqrt(np.mean(res.fun**2)))
    except Exception as e:
        print("model2 failed", e)
# try a rational with 2/2
def model3(p,x):
    a,b,c,d = p
    return (1+a*x+b*x*x)/(1+c*x+d*x*x)
def resid3(p):
    return model3(p,x)-y
for init in [np.array([0.0,0.0,0.3,0.0]), np.array([0.0,0.0,0.2,0.1]), np.array([-0.1,0.0,0.3,0.0])]:
    try:
        res=opt.least_squares(resid3, init, max_nfev=20000)
        print("model3", init, "->", res.x, "rmse", np.sqrt(np.mean(res.fun**2)))
    except Exception as e:
        print("model3 failed", e)
# inspect correlation and maybe monotonic
print("\nCorr:", train_df.corr(numeric_only=True)[target_col].to_string())
print("\nSorted values:")
print(train_df.sort_values(input_cols[0]).head(12).to_string(index=False))
print(train_df.sort_values(input_cols[0]).tail(12).to_string(index=False))

|B|TURN 1 -- HARNESS   <python_output>|B|
cols: ['Q2_GeV2'] target: GE_over_GD
 GE_over_GD  Q2_GeV2
      1.000    0.007
      0.996    0.012
      0.995    0.017
      0.993    0.022
      0.988    0.030
      0.987    0.038
      0.974    0.041
      0.981    0.048
      0.980    0.057
      0.990    0.061
Describe:
       GE_over_GD    Q2_GeV2
count   37.000000  37.000000
mean     0.960108   0.467568
std      0.036645   0.516130
min      0.844000   0.007000
50%      0.973000   0.273000
max      1.000000   1.830000
Ranges: 0.007 1.83 0.844 1.0
fit [0.  0.  0.7 0. ] -> [10.873 11.446 -0.561  0.750] rmse 0.012229365400954102
fit [-0.1  0.   0.7  0. ] -> [10.876 11.450 -0.561  0.750] rmse 0.012229365377535942
fit [0.  0.  0.6 0.1] -> [10.877 11.450 -0.561  0.750] rmse 0.012229365378633365
fit [-0.2  0.   0.6  0.1] -> [10.876 11.450 -0.561  0.750] rmse 0.012229365378083551
best (0.012229365377535942, array([10.876, 11.450, -0.561,  0.750]))
model2 [0.3 0.  0. ] -> [ 0.152 -0.165  0.076] rmse 0.014329317656398409
model2 [0.5 0.  0. ] -> [ 0.152 -0.165  0.076] rmse 0.014329317656398183
model3 [0.  0.  0.3 0. ] -> [ 33.18 -13.08  34.08 -12.58] rmse 0.012540593019447446
model3 [0.  0.  0.2 0.1] -> [ 33.18 -13.08  34.08 -12.58] rmse 0.012540593015895895
Corr: GE_over_GD    1.000000
Q2_GeV2      -0.906334
|EL|[sorted head(12) omitted -- same low-Q^2 rows as head(10) above; tail(12) shown, it reveals the high-Q^2 decline]|EL|
 GE_over_GD  Q2_GeV2
      0.967    0.584
      0.981    0.689
      0.916    0.853
      0.920    1.020
      0.936    1.330
      0.889    1.520
      0.844    1.740
      0.856    1.830

|B|TURN 2 -- AGENT   <python>|B|   |EL|-- sweep six physically-plausible forms; check each extrapolated to Q^2 = 5.85|EL|
import numpy as np, scipy.optimize as opt, pandas as pd
x = X_train[:,0]
y = y_train
cands = []
def score(name, p, pred):
    rmse = np.sqrt(np.mean((pred-y)**2))
    # simple extrapolation checks on [0,6]
    xt = np.array([0.0, 1.83, 3.0, 5.85])
    yt = pred_fn(p, xt)
    return rmse, yt
# Candidate 1: 1/(1+a x + b x^2 + c x^3)
def pred1(p, x):
    a,b,c = p
    return 1.0/(1.0 + a*x + b*x*x + c*x*x*x)
def fit1(init):
    res = opt.least_squares(lambda p: pred1(p,x)-y, np.array(init), max_nfev=20000)
    return res.x, np.sqrt(np.mean(res.fun**2))
for init in ([0.1,0.1,0.1],[0.05,0.05,0.01],[0.2,0.0,0.05],[0.0,0.1,0.05], [0.3,0.0,0.0]):
    try:
        p,rm=fit1(init)
        print("f1", init, p, rm, "check", pred1(p, np.array([0,1.83,5.85])))
    except Exception as e:
        print("f1 fail", init, e)
|EL|[... 57 lines: candidates f2 = (1+a x)/(1+b x+c x^2+d x^3), f3 = 1 - a x/(1+b x+c x^2),|EL|
|EL| f4 = 1 - a x - b x^2/(1+c x), f5 = exp(-(a x + b x^2)), f6 = (r+(1-r)exp(-a x))/(1+b x+c x^2),|EL|
|EL| each fitted from 3-5 inits in the same try/except pattern ...]|EL|

|B|TURN 2 -- HARNESS   <python_output>|B|
f1 [0.1, 0.1, 0.1] [ 0.152 -0.165  0.076] check [1.  0.840 |HL|0.088|HL|]
f2 [0.0, 0.1, 0.1, 0.1] [10.876 11.450 -0.561  0.750] check [1.  0.847 |HL|0.325|HL|]
f3 [0.1, 0.1, 0.1] [ 0.548 14.517 -6.362] check [1.  0.840 |HL|1.024|HL|]
f5 [0.1, 0.1] [0.0686 0.0070] check [1.  0.861 |HL|0.526|HL|]
f6 [0.7, 0.5, 0.0, 0.1] [ 0.9615 13.093 -0.036  0.060] check [1.  0.848 |HL|0.339|HL|]
|EL|-- every form reaches train RMSE ~ 0.012, but at Q^2 = 5.85 they spread from 0.088 to 1.02|EL|

|B|TURN 3 -- AGENT   <final_formula>|B|   |EL|-- submit f6: monotone, y(0)=1, saturating|EL|
"""Saturating exponential-over-quadratic fit for GE/GD versus Q^2."""
import numpy as np
USED_INPUTS = ["Q2_GeV2"]
LAW_CONSTANTS = {}
OTHER_CONSTANTS = {}
LOCAL_FITTABLE = {}
def predict(X):
    x = X[:, 0]
    r = 0.96150175
    a = 13.09262516
    b = -0.03621592
    c = 0.05985393
    return (r + (1.0 - r) * np.exp(-a * x)) / (1.0 + b * x + c * x * x)

|B|SCORE|B|  |N|S_N = 0.436  (third of 14 models; 8/14 score 0.000) ;  S_V = 1.0 ;  still below the 8-constant Arrington reference.|N|
\end{trajlisting}
\end{trajbox}

\subsection{\textsc{Real} $\times$ multi-group: CO$_2$ adsorption}
\label{app:case_co2}
\begin{trajbox}{\texttt{co2\_adsorption\_toth}, GPT-5.5 \quad(faithfully excerpted)}
\begin{trajlisting}
|B|TASK|B|       |N|co2_adsorption_toth :  amount of CO2 adsorbed  n  as a function of  pressure  P|N|
|B|DATA|B|       |HL|799 rows preloaded, spanning 42 clusters (group_id); scored on UNSEEN clusters|HL| |N|; P in [0.00013, 133.6]|N|
|B|SUBMIT|B|     |HL|non-empty LOCAL_FITTABLE (<=3 per-cluster params) + fit(X_fit,y_fit) + def predict(X, **params)|HL| |N|; ONE shared form|N|
|B|CAPS|B|       |N|max_law_constants=0 (no global fitted constants) ; fit() must run < 10 s per cluster|N|
|B|SCORED BY|B|  |HL|S_N -- RMSE on a held-out window of each test cluster, averaged over clusters|HL|

|B|TURN 1 -- AGENT   <python>|B|   |EL|-- inspect all 42 clusters, then fit standard isotherm families to each|EL|
import numpy as np, pandas as pd
from scipy.optimize import curve_fit
df=train_df.copy(); df['group']=group_ids_train
summ=df.groupby('group').agg(nrows=('P_bar','size'), Pmin=('P_bar','min'), Pmax=('P_bar','max'),
                             ymin=(target_col,'min'), ymax=(target_col,'max'))
print(summ.head().to_string())
def lang(P, q, b):     return q*b*P/(1+b*P)
def freundlich(P,K,m): return K*np.maximum(P,1e-12)**m
def sips(P,q,b,m):     P=np.maximum(P,0); z=np.maximum(b*P,1e-300)**m; return q*z/(1+z)
def toth(P,q,b,t):     P=np.maximum(P,0); z=np.maximum(b*P,1e-300); return q*z/((1+z**t)**(1/t))
forms = {'lang':(lang,[1,1],..), 'freund':(freundlich,[1,.5],..), 'sips':(sips,[1,1,1],..), 'toth':(toth,[1,1,1],..)}
for name,(func,p0,bounds) in forms.items():
    errs=[]
    for g in np.unique(group_ids_train):
        sub=df[df.group==g].sort_values('P_bar'); P=sub.P_bar.values; y=sub[target_col].values
        |EL|# set pinit[0] from y-max, pinit[1] from the half-saturation pressure|EL|
        popt,_=curve_fit(func,P,y,p0=pinit,bounds=bounds,maxfev=20000)
        errs.append(rmse(func(P,*popt),y))
    print(name, "mean_rmse", np.nanmean(errs), "median", np.nanmedian(errs))

|B|TURN 1 -- HARNESS   <python_output>|B|
group  nrows   Pmin     Pmax     ymin   ymax          |EL|# each cluster is one isotherm, saturating at high P|EL|
    0     13   0.019    0.962    2.90   6.22
    3     25   0.0058  30.04     0.31  10.9
lang    mean_rmse 0.2593   median 0.1882
freund  mean_rmse 0.3483   median 0.1967
sips    mean_rmse 0.1139   median 0.0422
toth    mean_rmse 0.1142   median 0.0371          |EL|# Sips / Toth clearly beat Langmuir|EL|
|EL|[... on its next attempts the agent hits three real guardrails, each rejected by the harness: ...]|EL|
Python execution failed: ValueError: operands could not be broadcast together with shapes (30,) (5,)
Python execution failed: TimeoutError: Python execution exceeded 100s
Python execution failed: ValidationError: Code appears to use high-dimensional brute-force search (4 nested for-loops). Use vectorized fitting.
|EL|-- it re-runs the Toth fit, vectorized and with fewer starts, until it passes|EL|

|B|TURN 6 -- AGENT   <final_formula>|B|   |EL|-- submit the Toth isotherm as the one shared per-cluster form|EL|
"""Three-parameter Toth adsorption isotherm shared across clusters: n = n_s*b*P / (1 + (b*P)^t)^(1/t)."""
import numpy as np
from scipy.optimize import least_squares
USED_INPUTS = ["P_bar"]
LAW_CONSTANTS = {}
OTHER_CONSTANTS = {}
|HL|LOCAL_FITTABLE = {"n_s":{"init":None}, "b":{"init":None}, "t":{"init":None}}|HL|
def _toth(P, n_s, b, t):                              |EL|# n = n_s * (b P) / (1 + (b P)^t)^(1/t)|EL|
    |EL|[... evaluated in log-space (logaddexp) for numerical stability ...]|EL|
    return out
|HL|def fit(X_fit, y_fit):|HL|                          |EL|# harness runs this on each UNSEEN cluster's window|EL|
    P = X_fit[:,0]; y = y_fit
    |EL|# log-space multi-start least_squares for (n_s, b, t) on this cluster|EL|
|EL|[... 30 lines: initial guesses from y-max and the half-saturation pressure, 3 starts, keep best-RMSE ...]|EL|
    return {"n_s": n_s, "b": b, "t": t}
|HL|def predict(X, n_s, b, t):|HL|
    return _toth(X[:,0], n_s, b, t)

|B|SCORE|B|  |N|S_N = 0.534  (7 of 14 models above the reference): the same Toth form transfers to unseen clusters, above the reference.|N|
\end{trajlisting}
\end{trajbox}

\subsection{\textsc{Parallel} $\times$ single-group: the Keeling curve}
\label{app:case_keeling}
\begin{trajbox}{\texttt{keeling\_curve}, GPT-5.5 \quad(faithfully excerpted)}
\begin{trajlisting}
|B|TASK|B|       |N|keeling_curve :  atmospheric CO2  c  as a function of  decimal year  t   (Mauna Loa)|N|
|B|DATA|B|       |HL|no rows preloaded -- the agent must call <experiment> to collect observations|HL| |N|; t in [1958, 2026]|N|
|B|SUBMIT|B|     |HL|def predict(X), constants baked in as literals, no fit(), no group_id|HL| |N|; <= 12 global constants|N|
|B|SCORED BY|B|  |HL|S_S -- does the submitted form match the hidden simulator's structure|HL|

|B|TURN 1 -- AGENT   <experiment>|B|   |HL|<- Parallel only: the agent designs its own measurements|HL|
{"year_decimal":[1958.2027,1958.2877,1958.3699,1958.4548,1958.5370,1958.6219,1958.7068,1958.7890,
                 1958.8740,1958.9562,1959.0411,1959.1260, 1960.5,1970.5,1980.5,1990.5,2000.5,2010.5,2020.5,2025.5],
 "n_samples":3, "seed":0}                    |EL|# 20 design points x 3 replicates -> the simulator adds noise|EL|

|B|TURN 1 -- HARNESS   <experiment_output>|B|
{"experiment_id": 1, "n_returned": 60, "preview_rows": [
  {"co2_ppm": 311.71, "year_decimal": 1958.2027}, {"co2_ppm": 315.19, "year_decimal": 1958.2027},   |EL|# replicates,|EL|
  {"co2_ppm": 312.28, "year_decimal": 1958.2027}, {"co2_ppm": 313.36, "year_decimal": 1958.2877}], ..}   |EL|# noisy|EL|
|EL|[... 6 more <experiment> rounds (msgs 8-29): denser time grids + replicate sampling, 7 experiment calls total ...]|EL|

|B|TURN k -- AGENT   <python>|B|   |EL|-- fit an exponential-accumulation trend (the hidden family) vs a plain quadratic|EL|
g=df.groupby('year_decimal')['co2_ppm'].agg(mean='mean', n='count')          |EL|# average the replicates|EL|
tg=g.year_decimal.values; yg=g['mean'].values; wg=np.sqrt(g.n.values)
def fit_exp(h=0):                                     |EL|# a + b*exp(c*(t-1958)) + h seasonal harmonics|EL|
    def pred(p,t):
        out = p[0] + p[1]*np.exp(p[2]*(t-1958.0))
        for k in range(1,h+1): out += p[idx]*np.sin(2*np.pi*k*t) + p[idx+1]*np.cos(2*np.pi*k*t)
        return out
    |EL|[... multi-start least_squares; report the weighted RMSE for h = 0..4 ...]|EL|
for h in [0,1,2,3,4]: fit_exp(h)
|EL|# also a plain quadratic in (t-2000), for comparison|EL|
x=tg-2000; cq=np.linalg.lstsq(np.column_stack([np.ones_like(x),x,x*x])*wg[:,None], yg*wg, rcond=None)[0]
print("quad", cq)

|B|TURN N -- AGENT   <final_formula>|B|   |EL|-- submits the QUADRATIC, not the exponential it just fit|EL|
"""Accelerating CO2 trend modeled as a quadratic function of decimal year."""
import numpy as np
USED_INPUTS = ["year_decimal"]
LAW_CONSTANTS = {}
OTHER_CONSTANTS = {}
LOCAL_FITTABLE = {}
def predict(X):
    t = X[:, 0]; x = t - 2000.0
    return 369.399866 + 1.88044598*x + 0.0137478424*x*x

|B|SCORE|B|  |N|S_S = 0.25: the quadratic tracks the data but is NOT the hidden exponential-accumulation family|N|
      |N|A + B*exp(k*(t-T0)).  On the SciLaws-Real version of this task the same model instead submits a cubic|N|
      |N|trend plus three annual harmonics and scores S_N = 0.78, beating the 12-constant NOAA reference.|N|
\end{trajlisting}
\end{trajbox}

\subsection{\textsc{Parallel} $\times$ multi-group: Sellmeier dispersion}
\label{app:case_sellmeier}
\begin{trajbox}{\texttt{optical\_dispersion\_sellmeier}, GPT-5.5 \quad(faithfully excerpted)}
\begin{trajlisting}
|B|TASK|B|       |N|optical_dispersion_sellmeier :  refractive index  n  as a function of  wavelength lambda|N|
|B|DATA|B|       |HL|no rows preloaded -- the agent must call <experiment> to collect its own observations|HL| |N|; 14 crystal clusters|N|
|B|SUBMIT|B|     |HL|non-empty LOCAL_FITTABLE (<=4 per-cluster params) + fit(X_fit,y_fit) + def predict(X, **params)|HL| |N|; ONE shared form|N|
|B|CAPS|B|       |N|max_law_constants=0 (no global fitted constants) ; <=10 experiment calls, <=20 points each|N|
|B|SCORED BY|B|  |HL|S_S -- does the submitted form match the hidden simulator's structure (up to algebraic equivalence)|HL|

|B|TURN 1 -- AGENT   <experiment>|B|   |HL|<- Parallel only: the agent designs its own measurements|HL|
{"wavelength_um":[0.4,0.5,1.0,2.0,2.6, 0.4,0.5,1.0,2.0,2.6, 0.8,1.0,2.0,4.0, 0.8,1.0,2.0,4.0],
 "group_id":[0,0,0,0,0, 1,1,1,1,1, 2,2,2,2, 3,3,3,3], "n_samples":1, "seed":0}

|B|TURN 1 -- HARNESS   <experiment_output>|B|
{"experiment_id": 1, "n_returned": 18, "n_appended_to_train": 18, "preview_rows": [
  {"n": 1.5467, "wavelength_um": 0.4, "group_id": 0}, {"n": 1.5379, "wavelength_um": 0.5, "group_id": 0},
  {"n": 1.5250, "wavelength_um": 1.0, "group_id": 0}, {"n": 1.5115, "wavelength_um": 2.0, "group_id": 0},
  {"n": 1.5369, "wavelength_um": 0.4, "group_id": 1}, {"n": 1.5033, "wavelength_um": 2.0, "group_id": 1}],
 |EL|[caps / row-id bookkeeping omitted]|EL| "n_clipped": 0}
|EL|[... 5 more <experiment> rounds (msgs 6-23): the agent refines the wavelength grid and adds clusters,|EL|
|EL| collecting 158 rows over all 14 groups, interleaved with the <python> analysis below ...]|EL|

|B|TURN k -- AGENT   <python>|B|   |EL|-- fit standard dispersion families to every cluster, compare max-error|EL|
def fit_group(df):
    lam=df.wavelength_um.values.astype(float); n=df.n.values.astype(float)
    # linear Cauchy n = A+B/l2+C/l4+D/l6
    for deg in [1,2,3]:
        Phi=np.vstack([lam**(-2*k) for k in range(0,deg+1)]).T
        coef=np.linalg.lstsq(Phi,n,rcond=None)[0]
        print(" n Cauchy deg",deg,"maxerr",np.max(np.abs(Phi@coef-n)))
    # generalized single pole n^2 = A + B/(l2-C)
    def f(l,A,B,C): return np.sqrt(A+B/(l*l-C))
    # Sellmeier n2=A+B*l2/(l2-C)
    def fs(l,A,B,C): return np.sqrt(A+B*l*l/(l*l-C))
|EL|[... then, after seeing the single-pole beat plain Cauchy, the agent adds an IR term (msg 14): ...]|EL|
# 1) n^2 = A + B/(lambda^2-C) + D*lambda^2 ? (Sellmeier plus IR term)   <- 4 params allowed
def f4(p):
    A,B,C,D=p; return np.sqrt(A + B/(l*l-C) + D*l*l) - n
|EL|[... fits f4 on every cluster; max-error drops to ~1e-4, best of all families tried ...]|EL|

|B|TURN N -- AGENT   <final_formula>|B|   |EL|-- submit the single-pole Sellmeier + IR term, fit per cluster|EL|
"""Single-resonance Sellmeier dispersion with a long-wavelength quadratic correction:
   n^2 = A + B/(lambda^2 - C) + D*lambda^2. Fit separately for each crystal cluster."""
import numpy as np
from scipy.optimize import least_squares
USED_INPUTS = ["wavelength_um"]
LAW_CONSTANTS = {}
OTHER_CONSTANTS = {}
|HL|LOCAL_FITTABLE = {"A":{"init":None}, "B":{"init":None}, "C":{"init":None}, "D":{"init":None}}|HL|
def _pred(lam, A, B, C, D):
    l2 = lam * lam
    return np.sqrt(np.maximum(A + B/(l2 - C) + D*l2, 1e-12))
|HL|def fit(X_fit, y_fit):|HL|                          |EL|# harness runs this on each UNSEEN cluster|EL|
    lam = np.asarray(X_fit[:,0], float); y = np.asarray(y_fit, float)
    |EL|# linear warm-start for (A,B,D) over a sweep of the pole C, then bounded least_squares|EL|
|EL|[... 20 lines: multi-start over C in {0, .01, .05, .15, .30, .60}*min(l2), keep best-RMSE ...]|EL|
    return {"A": A, "B": B, "C": C, "D": D}
|HL|def predict(X, A, B, C, D):|HL|
    return _pred(np.asarray(X[:,0], float), A, B, C, D)

|B|SCORE|B|  |N|S_S = 1.0  (missed = none): recovers the hidden n^2 = A + B/(l^2-C) + D*l^2 family exactly.|N|
      |N|On the SciLaws-Real version of this task the same model instead submits a Cauchy-type series and scores|N|
      |N|S_N = 0.21, below the strongest published baseline (the 4-term Cauchy equation).|N|
\end{trajlisting}
\end{trajbox}

\end{document}